\documentclass[twocolumn]{article}
\usepackage[english]{babel}
\usepackage[letterpaper,top=2cm,bottom=2cm,left=3cm,right=3cm,marginparwidth=1.75cm]{geometry}
\usepackage{amsmath}
\usepackage{graphicx}
\usepackage[colorlinks=true, allcolors=blue]{hyperref}
\usepackage{authblk}
\usepackage{hyperref}
\usepackage{cleveref}
\usepackage{listings}
\usepackage{microtype}
\usepackage{graphicx}
\usepackage{subfigure}
\usepackage{booktabs} 
\usepackage{soul}
\usepackage{ulem}
\usepackage{multirow}
\usepackage{verbatim}
\usepackage{empheq}
\usepackage{diagbox}
\usepackage{longtable}
\usepackage{listings}
\usepackage{amsmath}
\usepackage{amssymb}
\usepackage{mathtools}
\usepackage{amsthm}
\usepackage{blindtext}
\usepackage{titlesec}
\usepackage{array}
\newcolumntype{C}[1]{>{\centering}p{#1}}
\title{Unsupervised Learning of Hybrid Latent Dynamics:\\ A Learn-to-Identify Framework}
\author[1]{Yubo Ye}
\author[2]{Sumeet Vadhavkar}
\author[2]{Xiajun Jiang}
\author[2]{Ryan Missel}
\author[1]{Huafeng Liu}
\author[2]{Linwei Wang}
\affil[1]{Zhejiang University, China}
\affil[2]{College of Computing and Information Sciences, Rochester Institute of Technology, USA}

\begin{document}{
\maketitle

\begin{abstract}
Modern applications increasingly require \textit{unsupervised learning of latent dynamics} 
from high-dimensional time-series. 
This 
presents a significant challenge of \textit{identifiability}: many abstract latent representations may reconstruct observations, yet do they guarantee an adequate identification of the governing dynamics? 
This paper investigates this challenge 
from two angles: 
the use of 
physics inductive bias 
specific to the data being modeled, 
and a \textit{learn-to-identify} strategy that 
separates forecasting objectives from the data used for the identification. 
We combine these two strategies in a novel framework for \textit{unsupervised
meta-learning of hybrid latent dynamics} (Meta-HyLaD) with: 
1) 
a latent dynamic function
that hybridize known mathematical expressions of prior physics 
with neural functions describing its unknown errors, 
and 2) a meta-learning formulation to learn to separately identify both components of the hybrid dynamics. 
Through extensive experiments on five physics and one biomedical systems, 
we provide strong evidence for 
the benefits of Meta-HyLaD to integrate rich prior knowledge while identifying their gap to observed data. 
\end{abstract}

\footnote{It's a preprint under review}
\section{Introduction}

Learning the dynamics underlying observed time-series is at the heart of many applications, such as health monitoring 
and autonomous driving. 
As the quality and diversity of observation data continue to improve, 
modern applications increasingly require deep learning capabilities to 
extract \textit{latent} dynamics from high-dimensional observations (\textit{e.g.,} images), without label access to 
the 
latent variables being modeled. 
This \textit{unsupervised learning of latent dynamics} presents a fundamental challenge of \textit{identifiability} that is distinct from supervised 
modeling 
at the data space: 
different latent abstraction 
may be learned to reconstruct an observed time-series, but do they all guarantee 
an adequate identification of the governing dynamic functions? 

Recent works have shown that integrating inductive bias, in the form of known physics-based dynamic functions at the latent space, enables identification of the latent state variables and even parameters of the dynamic function with simple unsupervised objectives of data reconstruction, as illustrated in Fig. \ref{fig:overview}A  
 \cite{ALPS, gradsim, physics_inverer}. 
 These approaches however require the physics of the latent dynamic function to be accurately known, which is not always possible in practice. 
 \begin{figure*}
     \centering
     \includegraphics[width=.9\linewidth]{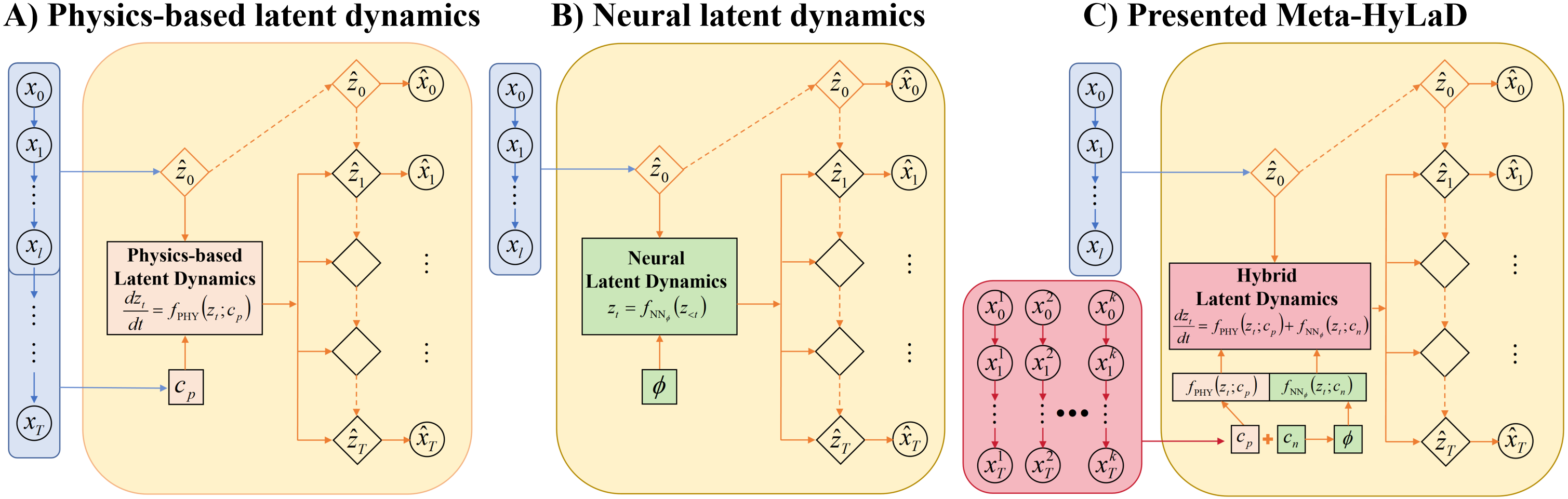}
     \caption{Overview of Meta-HyLaD \textit{vs}.\ existing frameworks for unsupervised learning of physics-based or neural latent dynamic functions.} 
     \label{fig:overview}
 \end{figure*}

This limitation has been recognized in emerging \textit{hybrid models} that combines physics-based dynamic functions with neural networks describing potentially unknown 
components 
in the prior physics \cite{hwangbo2019learning, zeng2020tossingbot, golemo2018sim,NeuralSim,UDE}. 
The existing hybrid models, however, are all learned at the data space with direct supervision on 
the state variables being modeled by 
the dynamic function. 
There are no existing solutions supporting 
\textit{unsupervised} learning of hybrid dynamic functions at the latent space. 

In this paper, we investigate 
the identifiability of \textit{neural} latent dynamic functions and 
show that -- unlike physics-based functions -- a \textit{neural} latent dynamic function without physics inductive bias 
cannot be adequately identified with an unsupervised \textit{reconstruction} objective.
Indeed, 
there has been a variety of 
purely neural-network based latent dynamic models which, 
as illustrated in \ref{fig:overview}B, 
are learned with an unsupervised \textit{forecasting} objective to use 
several initial frames to predict a longer sequence 
\cite{HGN, LNN, botev2021priors}. 
These latent dynamic functions, 
however, 
are optimized globally for the entire training data: 
\textit{i.e.,} 
the training process results in the identification of a single dynamic function without retraining.
Adopted directly in hybrid latent dynamics, this will result in a strong assumption that the gap between the prior physics and observed data are globally shared across all samples. 
This challenge 
may be further aggravated when the physics-based and neural components -- both heterogeneous across samples -- need to be separately identified.

}




To this end, we introduce the first solution to unsupervised learning of hybrid latent dynamics (Meta-HyLaD). 
As illustrated in Fig.~\ref{fig:overview}C, 
in contrast to existing purely 
physics-based 
or neural approaches 
(Fig.~\ref{fig:overview}A-B), 
Meta-HyLaD has two key innovations. 
First, 
inspired by \cite{UDE},
we formulate the latent dynamic function 
in the form of a universal differential equation (UDE), 
$\frac{d \mathbf{z}_t}{dt}=f(\mathbf{z}_t; \mathbf{c}_p, \mathbf{c}_n) = f_{\textrm{PHY}}(\mathbf{z}_{t};\mathbf{c}_p) + f_{\textrm{NN}_\phi}(\mathbf{z}_{t};\mathbf{c}_n)$, 
where $f_{\textrm{PHY}}(\mathbf{z}_{t},\mathbf{c}_p)$ describes prior physics specific to the data under study and $f_{\textrm{NN}_\phi}(\mathbf{z}_{t};\mathbf{c}_n)$ is a neural network describing potential errors in the prior physics,
each
with parameter $\mathbf{c}_p$ and 
$\mathbf{c}_n$ to be identified. 
Second, 
we present a novel \textit{learn-to-identify} solution 
where, 
instead of learning/identifying a single $f_{\textrm{NN}_\phi}$ that can only describe a \textit{gloal} discrepancy between $f_{\textrm{PHY}}$ and the true governing dynamics, 
we 
train feedforward meta-models to learn to identify 
$f_{\textrm{NN}_\phi}$: 
we will show that this strategy -- 
asking latent dynamic functions to forecast for samples different from those used to identify them -- 
is critical to their successful identifications.

We evaluate Meta-HyLaD on a spectrum of dynamic systems ranging from relatively simple benchmark physics
to tracer kinetics underlying dynamic positron emission tomography (PET) imaging \cite{PET}.
We first demonstrate that the unsupervised reconstruction objectives cannot guarantee the adequate identification of a neural latent dynamic function, 
and that this can be overcome either by the incorporation of a physics-based dynamic function or -- at the absence of such physics inductive bias -- 
the presented learn-to-identify formulation (Section \ref{subsec:exp:id}). 
We then demonstrate the clear improvements in system identification and time-series forecasting obtained by Meta-HyLaD in comparison to 
purely physics-based,  
purely neural, 
or hybrid models with a global neural component (Section \ref{subsec:exp:hybrid}). 
Finally, 
we demonstrate the advantages of Meta-HyLaD in comparison to a wide variety of existing works -- both physics- and neural-based latent dynamic models and their extensions to meta-learning formulations -- across different types of 
errors in the prior physics and across the datasets considered (Section \ref{subsec:exp:baselines}). 
The results provide strong evidence for the benefits of hybrid dynamics to integrate rich prior knowledge while allowing for their errors, as well as the importance of properly identifying these hybrid components. 




\section{Related Works}



\textbf{Unsupervised learning of latent dynamics:} There has been a surge of interests in learning the governing dynamics of high-dimensional time-series, 
mostly in an encoder/decoder formulation 
with the goal to identify a dynamic function  and its  
resulting state variables at the latent space. 

\textbf{\textit{Physics-based (white-box) latent dynamics:}}
Several methods have proposed to integrate physics-based dynamic equations 
into the latent space as prior knowledge specific to the data under study \cite{ALPS, gradsim, physics_inverer}. 
They however require the physics of the latent dynamics to be accurately known (with only some physics parameters unknown), which is not always possible in practice. 
We will show that -- at the presence of different types of errors in the prior physics -- the performance of this type of \textit{white-box} modeling quickly deteriorates. 
How to leverage physics-based functions while addressing their errors remains an open question in learning latent dynamics.


\textbf{\textit{Neural (black-box) latent dynamics:}}
There has also been an explosion of neural-network based latent dynamic models, 
using for instance LSTMs \cite{VRNN, DKF,KVAE, DVBF} and  
neural ODEs \cite{ODE2VAE, botev2021priors}  
at the latent space. 
Different from \textit{white-box} dynamic functions with physically-meaningful latent state variables and parameters, 
\textit{black-box} dynamic functions are associated with abstract latent states
and functions. It thus sees an increased challenge of \textit{identifiability},
especially 
when learning 
across heterogeneous dynamics.
Among recent efforts,
there has been a growing body of 
works exploring ways to effectively identify different neural dynamic functions from  heterogeneous data samples
\cite{LEADS,DyAd,CoDA,jiang2022sequential}, among which 
meta-learning has emerged to be a promising solution 
\cite{CoDA,jiang2022sequential}. 

All of these existing works, however, 
are based on \textit{black-box} dynamic functions without the ability to leverage potentially valuable prior knowledge. 
Some recent works, 
such as Hamiltonian and Lagrangian neural network \cite{HNN&LNN1, HNN&LNN2, HNN&LNN3, HNN&LNN4, HNN&LNN5, HGN, HNN&LNN7, HNN&LNN8, HNN&LNN6, HNN&LNN9, HNN&LNN10}
encode 
physical laws as priors to constrain the behavior of the neural latent dynamic function, 
blurring the boundary between black- and white-box modeling. 
The latent dynamic functions, however, 
are still in the form of neural networks. 
Furthermore, these approaches 
encompass broader priors rather than richer knowledge specific to the problem/data. 

Meta-HyLaD represents the first hybrid solution to learning latent dynamics. 
Moreover, 
we dive deep into the \textit{identifiability} of physics-based \textit{versus} neural components,  
and show that -- for latent dynamic functions without physics inductive bias --  successful reconstruction of observed time-series does not guarantee a successful identification: 
a non-trivial challenge inadequately discussed in the literature. 

\textbf{Hybrid (grey-box) dynamics:} 
When the dynamic function is directly supervised, 
a number of hybrid models 
has emerged to
combine physics-based functions with neural networks to compensate for unknown components in the known physics. 
Most approaches \cite{hwangbo2019learning, zeng2020tossingbot, golemo2018sim} use a neural function to describe the residual between the simulated and the measured states.
NeuralSim \cite{NeuralSim} takes it a step further and 
includes neural functions as different components within a physics-based function.
UDEs define a differential equation with known mathematical expressions combined with 
neural networks \cite{UDE}. 

These hybrid models, however, are learned at the data space and directly supervised by measured state variables of the dynamics. 
The absence of such supervision
substantially increases the difficulty to \textit{identify} the dynamic function, and 
Meta-HyLaD represents the first step towards a solution.

\section{Methods}
\label{sec:method}

Fig.~\ref{fig:overview}C outlines the key elements in Meta-HyLaD:
1) hybrid latent dynamic functions, 
and 2) learn-to-identify strategies. 


\subsection{Hybrid Latent Dynamic Functions as a UDE}


Considering high-dimensional time-series $\mathbf{x}_{1:T} = [ \mathbf{x}_0, 
\cdots, \mathbf{x}_t \cdots, \mathbf{x}_T ]$, we model its generation process as: 
\begin{equation}
\label{eq:SSM}
\mathbf{z}_t=\mathcal{F}(\mathbf{z}_{<t}); \quad 
\mathbf{x}_t=g(\mathbf{z_t})
\end{equation}
where function $\mathcal{F}(\mathbf{z}_{<z})$ describes the latent dynamics of the state variables $\mathbf{z}_t$, 
and function $g(\mathbf{z_t})$ describes the emission of the latent state variables to the observed data. 

\textbf{Latent dynamic functions:} While Meta-HyLaD is agnostic to the type of dynamic functions used in Equation (\ref{eq:SSM}), 
we choose an ordinary differential equation (ODE) to describe the latent dynamics as a \textit{continuous} process that can be emitted to the observation space only when needed. 
More specifically, 
we describe the latent dynamics as a UDE:
\begin{equation}
\label{eqn:hybrid}
    \frac{d\mathbf{z}_t}{dt}=f_\phi(\mathbf{z}_t;\mathbf{c}_\mathbf{z})=f_{\textrm{PHY}}(\mathbf{z}_t; \mathbf{c}_p)+f_{\textrm{NN}_\phi}(\mathbf{z}_t; \mathbf{c}_n)
\end{equation}
where $f_\textrm{PHY}(\mathbf{z}_t;\mathbf{c}_p)$ represents the known physics equation governing the data, and
$\mathbf{c}_p$ its unknown 
parameters.
$f_{\textrm{NN}_\phi}(\mathbf{z}_t;\mathbf{c}_n)$ represents the potential errors in the prior physics, 
modeled by a neural network with weight parameters $\phi$. 
Instead of identifying a single neural function 
$f_{\textrm{NN}_\phi}(\mathbf{z}_t)$ that will only model a global discrepancy between $f_\textrm{PHY}$ and the true governing dynamics, 
we further allow it to change with a \textit{parameter} $\mathbf{c}_n$ that can be identified from data: 
more specifically,  
we use $\mathbf{c}_n$ to generate the weight parameter $\phi$ of $f_{\textrm{NN}_\phi}$ via a hyper network $\phi = h_\theta(\mathbf{c}_n)$. 




\textbf{Emission functions:} 
The emission function 
$\mathbf{x}_t=g(\mathbf{z_t})$, 
\textit{i.e.,} the decoder, 
bridges the mapping from the latent state space to the observed data space. 
In most existing works considering latent physics-based functions, it has been considered critical 
for this emission function to be physics-based \cite{gradsim,physics_inverer}. 
Similarly, 
when the latent dynamic function is purely neural, 
it is customary to adopt a neural function as the emission function \cite{botev2021priors}. 
As a secondary objective of this paper, we will investigate the use of a physics-based \textit{vs}.\ neural decoder for different types of latent dynamic functions. 
We denote them generally as $g_\psi$ where $\psi$ is known if $g$ is physics-based, and unknown if $g$ is a neural network parameterized with $\psi$. 

With the pair of functions described in Equations (\ref{eq:SSM}-\ref{eqn:hybrid}),
both state variables $\mathbf{z}_{0:T}$ and data $\mathbf{x}_{0:T}$ can be generated once 
the initial condition $\mathbf{z}_{0}$ 
and the parameters of the hybrid components -- $\mathbf{c}_p$ and $\mathbf{c}_n$ -- are properly identified.



\subsection{Learn-to-Identify via Meta-Learning}
\label{subsec:ID}




Given an observed time-series $\mathbf{x}_{0:T} = [ \mathbf{x}_0, \cdots, \mathbf{x}_t \cdots, \mathbf{x}_T ]$ in training data $\mathcal{D}$, 
a common approach to identifying its latent dynamic process is 
to infer its time-varying latent state $\mathbf{z}_{0:T} = [ \mathbf{z}_0, \cdots, \mathbf{z}_t \cdots, \mathbf{z}_T ]$ to reconstruct $\mathbf{x}_{0:T}$ \cite{DKF,DVBF,KVAE,ALPS}. 
We will show that -- for neural functions with abstract latent states $\mathbf{z}_t$  -- 
this reconstruction 
does not guarantee a correct identification of
the dynamic function governing $\mathbf{z}_t$: an evidence is when the identified dynamic function, 
given different initial conditions,
could not generate the correct time series governed by the same dynamics.  

We propose that a proper identification of the latent dynamic functions requires us to go a step further: 
one must further separate $\mathbf{z}_{0:T}$ into two key ingredients generating it -- the initial latent state $\mathbf{z}_{0}$ that is specific to the time-series sample, and parameters $\mathbf{c}_\mathbf{z}$ for the governing latent dynamic function. 
In a neural network where the learned functions can be arbitrary and the latent states abstract without direct supervision, 
this separation is not trivial. 
Below, we describe our learning strategies to facilitate this separation, 
and discuss how such identification strategies may be different for the physics-based \textit{versus} neural functions. 

\textbf{Identifying initial latent states:} The initial latent state $\mathbf{z}_0$ of an observed time series $\mathbf{x}_{0:T}$  is specific to that series and can be simply identified from the first several frames of observations $\mathbf{x}_{0:l}$, where $l\ll T$:
\begin{equation}
\label{eqn:z0}
   \hat{\mathbf{z}}_0 = \mathcal{E}_{\phi_z} (\mathbf{x}_{0:l})
\end{equation}
 where $\mathcal{E}_{\phi_z}$ is a neural encoder with weight parameters $\phi_z$.

\begin{figure}[t]
    \centering
\includegraphics[width=.9\linewidth]{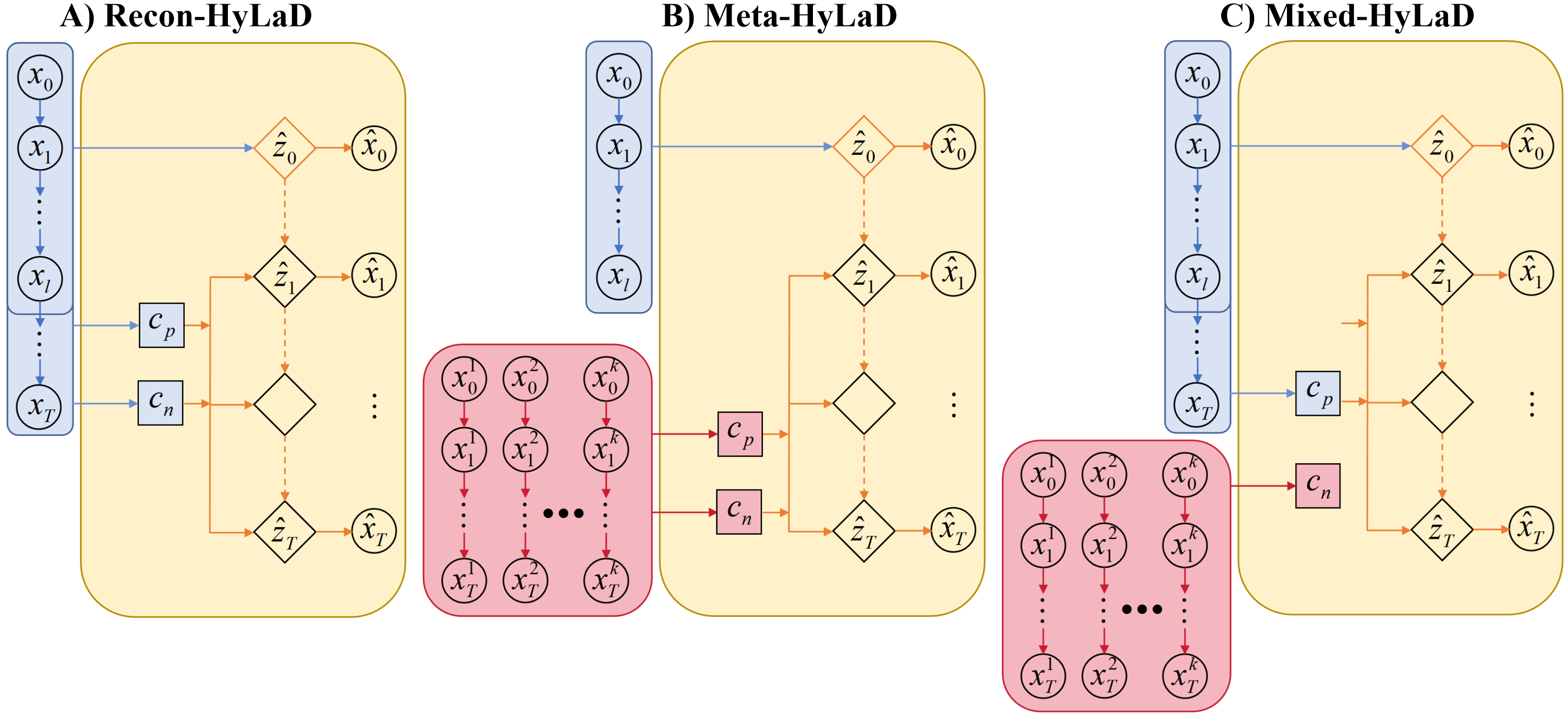}
\caption{HyLaD with alternative identification strategies}
    \label{fig:models}
\end{figure}

\paragraph{Identifying latent dynamics:} 
To identify $\mathbf{c}_n$ and $\mathbf{c}_p$ for Equation (\ref{eqn:hybrid}), 
a simple reconstruction objective would be: 
\begin{equation}
\label{eqn:cz_recon}
    \hat{\mathbf{c}}_\mathbf{z}=[\hat{\mathbf{c}}_p,\hat{\mathbf{c}}_n]=\mathcal{E}_{\phi_c} (\mathbf{x}_{0:T})
\end{equation}
\begin{equation}
\label{eqn:odeforward}
\hat{\mathbf{z}}_{t+1}=\hat{\mathbf{z}}_t+
    \int_{t}^{t+1} f_\phi(\hat{\mathbf{z}}_t;\hat{\mathbf{c}}_\mathbf{z})d\mathbf{z}, \quad t = 0,\cdots,T-1
\end{equation}
\begin{equation}
\label{eqn:obj_recon}
    \begin{aligned}
\hat{\Theta}
=\arg \min_{\Theta}
\frac{1}{\left | \mathcal{D} \right | }
\sum_{\mathbf{x}_{0:T}\in \mathcal{D}}\left \| \mathbf{x}_{0:T}-g_{\psi}(\hat{\mathbf{z}}_{0:T}) \right \|\|_2^2
    \end{aligned}
\end{equation}
where $\Theta= \left \{\phi_z, \phi_c \right \}$ and $\psi$ if $g_\psi$ is neural. 
$\mathcal{E}_{\phi_{c}}$ is neural encoder parameterized by $\phi_{c}$. 
We refer to this as a \textit{reconstruction objective}, as illustrated in Fig.~\ref{fig:models}A. 
Although commonly used, 
we will show that this learning strategy -- while deceivingly able to provide good reconstruction performance on a given time series -- 
will not be able to separately identify the three components of the latent dynamic function.

\textbf{\textit{Identifying the neural component:}} 
Instead of the reconstruction of an observed time-series, 
we present a novel \textit{learn-to-identify} solution to address the identifiability challenge 
associated with the neural component 
$f_{\textrm{NN}_\phi}$ of the latent dynamic function. 
The fundamental intuition is that, 
to separately identify time-varying latent states 
from time-invariant 
governing dynamic functions, 
one can leverage the statistical strength that 
the governing latent dynamics might be shared across multiple samples. 
Therefore, 
we can attempt to learn to identify $\mathbf{c}_n$ for $f_{\textrm{NN}_\phi}$ from one or more samples, and ask the identified function to be applicable for \textit{disjoint} samples sharing the same dynamics. 
This is the basis for the presented 
\textit{learn-to-identify} framework. 

Formally, we cast this into a meta-learning formulation. 
Consider a dataset $\mathcal{D}$ of high-dimensional time-series with $M$ similar but distinct underlying dynamics: \small{$\mathcal{D}=\left \{ \mathcal{D}_j\right \} _{j=1}^M$.} \normalsize For each $\mathcal{D}_j$, we consider disjoint few-shot context time-series samples \small{$\mathcal{D}_j^s= \left \{ \mathbf{x}_{0:T}^{s,1},\mathbf{x}_{0:T}^{s,2},\ldots,\mathbf{x}_{0:T}^{s,k} \right \}$} \normalsize and query samples \small{$\mathcal{D}_j^q= \left \{ \mathbf{x}_{0:T}^{q,1},\mathbf{x}_{0:T}^{q,2},\ldots,\mathbf{x}_{0:T}^{q,d} \right \}$ where $k\ll d$.} \normalsize 
Instead of the reconstruction objective in Equations (\ref{eqn:cz_recon}-\ref{eqn:obj_recon}), 
we formulate a meta-objective to learn 
to identify 
$\mathbf{c}_n$ 
from $k$-shot context time-series samples $\mathcal{D}_j^s$, 
such that the identified hybrid dynamic function is able to forecast for any query time series in $\mathcal{D}_j^q$
given only an estimate of its initial state $\hat{\mathbf{z}}_{0,j}^q$. 
More specifically, 
we have a feedforward meta-model
$\mathcal{M}_{\zeta_n}(\mathcal{D}_j^s)$
to learn to identify $\mathbf{c}_{n,j}$ 
for dynamics $j$ as:
\begin{equation}
\label{eqn:cn}
    \hat{\mathbf{c}}_{n,j} = \mathcal{M}_{\zeta_n}(\mathcal{D}_j^s) = \frac{1}{k}{\textstyle \sum_{\mathbf{x}_{0:T}^s\in \mathcal{D}_j^s}} \mathcal{E}_{\zeta_n} (\mathbf{x}_{0:T}^{s})
\end{equation}
where an embedding is extracted from each individual context time-series via a meta-encoder $\mathcal{E}_{\zeta_n}$ and gets
aggregated across $\mathcal{D}_j^s$ 
to 
 extract knowledge shared by the set. 
$k$ is the size of the context set, and its value can be fixed or variable which we will demonstrate in the ablation study. 


\textbf{\textit{Identifying the physics-component:}} 
The parameter $\mathbf{c}_p$ for the physics-based $f_{\textrm{PHY}}(\mathbf{z}_t; \mathbf{c}_p)$ of the latent dynamic functions can be inferred in a similar fashion. 
However, 
as 
the structural form 
of the function 
defines specific physics meaning for the latent state variable $\mathbf{z}_t$, 
we will show that 
this physics inductive bias alone 
allows it to be identified with a simple reconstruction objective 
similar to Equations (\ref{eqn:cz_recon}-\ref{eqn:obj_recon}). 
In another word, 
$\mathbf{c}_p$ for the physics-based dynamic function can be identified in two alternative formulations: 
\begin{empheq}[left={\hat{\mathbf{c}}_{p,j} =}\empheqlbrace]{align}
    \label{eqn:cp_meta}
   & \frac{1}{k}{\textstyle \sum_{\mathbf{x}_{0:T}^s\in \mathcal{D}_j^s}} \mathcal{E}_{\zeta_p} (\mathbf{x}_{0:T}^{s}),& 
    \text{Fig.~\ref{fig:models}B} \\
    \label{eqn:cp_recon}
    &\mathcal{E}_{\zeta_p} (\mathbf{x}_{0:T}^q),\quad \mathbf{x}_{0:T}^q \in \mathcal{D}_j^q   &\text{Fig.~\ref{fig:models}C}
\end{empheq}

\begin{table*}[t]
\caption{Summary of experimental settings.}
\label{table:systems}
\centering
\resizebox{\linewidth}{!}{
\begin{tabular}{cllll}
\hline
DataSet & \multicolumn{1}{c}{Physics} & \multicolumn{1}{c}{Equation} & \multicolumn{1}{c}{Configurantion} &\multicolumn{1}{c}{DataSize} \\ 
\hline
\multirow{2}{*}{Pendulum} & Full & $\frac{\mathrm{d}}{\mathrm{d} t}\begin{bmatrix} \varphi  \\ \dot{\varphi}  \end{bmatrix}= \begin{bmatrix} \dot{\varphi}  \\ -\frac{\color{blue}{G}}{L}\sin(\varphi)-{\color{red}{\beta}} \dot{\varphi} \end{bmatrix}$ & 
\multirow{2}{*}{
\begin{tabular}[c]{@{}l@{}} ${\color{blue}{G}}\in[5.0, 15.0]$
\\${\color{red}{\beta}}\in(0.0, 1.0]$ ID
\\${\color{red}{\beta}}\in(1.0, 1.2]$ OOD
\end{tabular}
} 
& \multirow{2}{*}{15390} \\ 
& Partial & $\frac{\mathrm{d}}{\mathrm{d} t}\begin{bmatrix} \varphi  \\ \dot{\varphi}  \end{bmatrix}= \begin{bmatrix} \dot{\varphi}  \\ -\frac{{\color{blue}{G}}}{L}\sin(\varphi) \end{bmatrix}$ & & \\ 
\hline
\multirow{2}{*}{
\begin{tabular}[c]{@{}c@{}}Mass Spring\\ 
$\Tilde{x}=x_1-x_2$ \\ $\Tilde{v}=v_1-v_2$\end{tabular}
}
& Full & $\frac{\mathrm{d}}{\mathrm{d} t}\begin{bmatrix} v_1  \\ v_2  \end{bmatrix}=
\begin{bmatrix} -\frac{\Tilde{x}}{\left | \Tilde{x} \right |} \frac{{\color{blue}{k}}}{m_1}
\left ( \left | \Tilde{x} \right | -l_0 \right ) - \frac{{\color{red}{\beta}}}{m_1} \Tilde{v} \\ \frac{\Tilde{x}}{\left | \Tilde{x} \right |} \frac{{\color{blue}{k}}}{m_2}
\left ( \left | \Tilde{x} \right | -l_0 \right ) +\frac{{\color{red}{\beta}}}{m_2} \Tilde{v}
\end{bmatrix}$ 
& \multirow{2}{*}{\begin{tabular}[c]{@{}l@{}} ${\color{blue}{k}}\in[5.0, 15.0]$
\\${\color{red}{\beta}}\in(0.0, 1.0]$ ID
\\${\color{red}{\beta}}\in(1.0, 1.5]$ OOD
\end{tabular}
} 
& \multirow{2}{*}{12960} \\
& Partial & $\frac{\mathrm{d}}{\mathrm{d} t}\begin{bmatrix} v_1  \\ v_2  \end{bmatrix}=
\begin{bmatrix} -\frac{\Tilde{x}}{\left | \Tilde{x} \right |} \frac{{\color{blue}{k}}}{m_1}
\left ( \left | \Tilde{x} \right | -l_0 \right ) \\ \frac{\Tilde{x}}{\left | \Tilde{x} \right |} \frac{{\color{blue}{k}}}{m_2} \left ( \left | \Tilde{x} \right | -l_0 \right )
\end{bmatrix}$
& & \\ 
\hline
\multirow{2}{*}{\begin{tabular}[c]{@{}c@{}}Bouncing Ball \\ \end{tabular}}
& Full & $\frac{\mathrm{d}}{\mathrm{d} t}\begin{bmatrix} v_x  \\ v_y  \end{bmatrix}=\begin{bmatrix}{\color{blue}{G}} \cos {\color{red}{\varphi}} \\{\color{blue}{G}}\sin {\color{red}{\varphi}}  \end{bmatrix}, {\color{red}{\varphi}}=\frac{{\color{red}{i}}}{8}\pi,i=-8,-7,\dots,6,7$ 
& \multirow{2}{*}{\begin{tabular}[c]{@{}l@{}} ${\color{blue}{G}}\in[0.0, 5.0]$
\\${\color{red}{i}}\in [-6:-1]\cup[2:7]$ ID
\\${\color{red}{i}}\in[-8,-7,0,1]$ OOD
\end{tabular}}  
& \multirow{2}{*}{21060} \\ 
& Partial & $\frac{\mathrm{d}}{\mathrm{d} t}\begin{bmatrix} v_x  \\ v_y  \end{bmatrix}=\begin{bmatrix} 0 \\-{\color{blue}{G}} \end{bmatrix}$ & & \\ \hline
\multirow{3}{*}{Rigid Body} & & $\begin{bmatrix} m & 0\\ 0 & \mathbf{I} \end{bmatrix}
\begin{bmatrix}\dot{\mathbf{v}}\\ \dot{\mathbf{\omega}} \end{bmatrix} = 
\begin{bmatrix}\mathbf{f}\\  \mathbf{r}\times\mathbf{f} \end{bmatrix} -
\begin{bmatrix}0\\ \mathbf{\omega}\times \mathbf{I} \mathbf{\omega}\end{bmatrix}$   
& \multirow{3}{*}{\begin{tabular}[c]{@{}l@{}} ${\color{blue}{\psi}}\in [-\pi, \pi)$
\\${\color{red}{\varphi}}\in (\frac{\pi}{8}, \frac{\pi}{2}]$ ID
\\${\color{red}{\varphi}}\in [0, \frac{\pi}{8}]$ OOD
\end{tabular}}  
& \multirow{3}{*}{4800} \\ 
& Full  & $\mathbf{f} =f\cdot [\cos{\color{blue}{\psi}}\sin{\color{red}{\varphi}} , \sin{\color{blue}{\psi}}\sin{\color{red}{\varphi}}, \cos{\color{red}{\varphi}}]$   
&  & \\
& Partial  & $\mathbf{f} =f\cdot [\cos{\color{blue}{\psi}}, \sin{\color{blue}{\psi}}, 0]$   &  &  \\ 
\hline
\multirow{2}{*}{\begin{tabular}[c]{@{}c@{}}Double Pendulum \\ 
$\Tilde{m}=m_1/m_2$ \\
$\phi_1=\cos(\varphi_1 - \varphi_2)$ \\ $\phi_2=\sin(\varphi_1 - \varphi_2)$
\end{tabular}}
& Full & $\frac{d}{dt}\begin{bmatrix} \dot{\varphi}_1  \\ \dot{\varphi}_2 \end{bmatrix}=
\begin{bmatrix}\frac{{\color{blue}{G}}\sin\varphi_2\phi_1 -\phi_2(L_1\dot{\varphi}_1^2\phi_1+L_2\dot{\varphi}_2^2)
-({\color{red}{\Tilde{m}}}+1){\color{blue}{G}}\sin\varphi_1}{L_1({\color{red}{\Tilde{m}}}+\phi_2^2)} \\
\frac{({\color{red}{\Tilde{m}}}+1)(L_1\dot{\varphi}_1^2\phi_2 -{\color{blue}{G}}\sin\varphi_2+{\color{blue}{G}}\sin\varphi_1\phi_1)
+L_2\dot{\varphi}_2^2\phi_1 \phi_2}{L_2({\color{red}{\Tilde{m}}}+\phi_2^2)}\end{bmatrix}$ 
& \multirow{2}{*}{\begin{tabular}[c]{@{}l@{}} ${\color{blue}{G}}\in[5.0, 15.0]$
\\${\color{red}{\Tilde{m}}}\in [0.5, 1.5]$ ID
\\${\color{red}{\Tilde{m}}}\in (1.5, 2.0]$ OOD
\end{tabular}}
& \multirow{2}{*}{12705} \\
& Partial & 
$\frac{d}{dt}\begin{bmatrix} \dot{\varphi}_1  \\ \dot{\varphi}_2 \end{bmatrix}=
\begin{bmatrix}\frac{{\color{blue}{G}}\sin\varphi_2\phi_1 -\phi_2(L_1\dot{\varphi}_1^2\phi_1+L_2\dot{\varphi}_2^2)
-2{\color{blue}{G}}\sin\varphi_1}{L_1(1+\phi_2^2)} \\
\frac{2(L_1\dot{\varphi}_1^2\phi_2 -{\color{blue}{G}}\sin\varphi_2+{\color{blue}{G}}\sin\varphi_1\phi_1)
+L_2\dot{\varphi}_2^2\phi_1 \phi_2}{L_2(1+\phi_2^2)}\end{bmatrix}$ & & \\ 
\hline
\multirow{2}{*}{Dynamic PET} & Full & 
$\begin{bmatrix}\dot{C}_{Ei}(t) \\ \dot{C}_{Mi}(t) \end{bmatrix}=
\begin{bmatrix} \color{red}{-k_2-k_3} & \color{red}{k_4}\\ \color{red}{k_3} & \color{red}{-k_4} \end{bmatrix}
\begin{bmatrix}C_{Ei}(t) \\ C_{Mi}(t)\end{bmatrix}+\begin{bmatrix}{\color{blue}{k_1}} \\ 0 \end{bmatrix}C_P(t)$ & \multirow{2}{*}{See Appendix \ref{appendix:data-PET}} & \multirow{2}{*}{2000} \\
& Partial & $\dot{C}_{Ti}(t)=-{\color{blue}{k_2}}C_{Ti}(t)+{\color{blue}{k_1}}C_P(t)$ & & \\ \hline
\end{tabular}
}
\end{table*}

\textbf{Learn-to-identify meta-objectives:} 
Given data with $M$ different dynamics $\mathcal{D}=\left \{ \mathcal{D}_j\right \} _{j=1}^M$, for 
all query samples $\mathbf{x}_{0:T}^{q} \in \mathcal{D}_j^q$, 
we have its initial latent state $\hat{\mathbf{z}}_0^q$ identified from its own initial frames $\mathbf{x}_{0:l}^{q}$ (Equation (\ref{eqn:z0})), 
and its hybrid latent dynamic functions identified with $\hat{\mathbf{c}}_{n,j}$ and $\hat{\mathbf{c}}_{p,j}$ as described in Equations (\ref{eqn:cn}-\ref{eqn:cp_meta}) 
(Fig.~\ref{fig:models}B, \textit{Meta-HyLaD}), 
or in Equations (\ref{eqn:cn}) and (\ref{eqn:cp_recon}) 
(Fig.~\ref{fig:models}C, \textit{Mixed-HyLaD}).
Given the inferred 
$\hat{\mathbf{z}}_0^q, \hat{\mathbf{c}}_{n,j}$, and $\hat{\mathbf{c}}_{p,j}$, 
we 
minimize the forecasting accuracy on the query time-series with $\Theta= \left \{\phi_z, \zeta_p, \zeta_n \right \}$ and 
$\psi$ if $g_\psi$ is neural:
\begin{equation}
\hat{\Theta} = 
\arg \min_\Theta  \sum_{j=1}^M \sum_{\mathbf{x}_{0:T}^q\in \mathcal{D}_j^q} \left \| \mathbf{x}_{0:T}^q-\hat{\mathbf{x}}_{0:T}^q \right \|_2^2 
\end{equation}

\section{Experiments and Results}
\label{sec:exp}

We first investigate the two key contributions of Meta-HyLaD:
1) 
its identification strategy in comparison to reconstruction objectives in addressing the 
identifiability of physics-based \textit{vs.} neural latent dynamic functions (Section \ref{subsec:exp:id}), and 2)
its modeling strategy \textit{vs.} purely physics-based, 
purely neural, or hybrid models with a global neural component 
(Section \ref{subsec:exp:hybrid}).  
We then evaluate Meta-HyLaD with a variety of existing physics-based and neural latent dynamic models \cite{ALPS,botev2021priors} along with their extensions into the presented meta-learning frameworks, 
considering both physics-based and neural decoders 
(Section \ref{subsec:exp:baselines}).
Finally, 
we test its feasibility  on identifying radiotracer kinetics in dynamic PET (Section \ref{subsec:exp:pet}).
All experiments are repeated with three random seeds.

\begin{table*}[t]
\caption{Comparison of HyLAD obtained from
different identification strategies in reconstruction and prediction performance.} 
\label{Experiment1_table}
\resizebox{\linewidth}{!}{
\begin{tabular}{|l|cccc|cccc|}
\hline
& \multicolumn{4}{c|}{Recontruction Task} & \multicolumn{4}{c|}{Prediction Task}\\ \hline
\diagbox{Strategy}{MSE} 
& \multicolumn{1}{c|}{$\mathbf{x}_t(e^{-4})\downarrow$} & \multicolumn{1}{c|}{$\mathbf{z}_t(e^{-3})\uparrow$} & \multicolumn{1}{c|}{$\mathbf{c}_p(e^{-1})\downarrow$} & \multicolumn{1}{c|}{$\mathbf{c}_n(e^{-3})\downarrow$} 
& \multicolumn{1}{c|}{$\mathbf{x}_t$($e^{-4}$)} & \multicolumn{1}{c|}{$\mathbf{z}_t$($e^{-3}$)} 
& \multicolumn{1}{c|}{$\mathbf{c}_p$($e^{-1}$)}  & \multicolumn{1}{c|}{$\mathbf{c}_n$($e^{-3}$)} \\ \hline
A -- Recon-HyLaD  & \multicolumn{1}{c|}{\textbf{3.95(0.27)}} & \multicolumn{1}{c|}{\textbf{4.34(0.74)}} & \multicolumn{1}{c|}{2.05(0.31)} & \textbf{1.13(0.26)} & \multicolumn{1}{c|}{7.17(0.50)} & \multicolumn{1}{c|}{10.59(1.06)} & \multicolumn{1}{c|}{1.83(0.15)} & 3.72(0.69) \\ \hline
B -- Meta-HyLaD (k=1) & \multicolumn{1}{c|}{4.39(0.19)} & \multicolumn{1}{c|}{5.19(0.85)} & \multicolumn{1}{c|}{\textbf{1.70(0.10)}} & 1.16(0.18) & \multicolumn{1}{c|}{4.40(0.37)} & \multicolumn{1}{c|}{6.70(1.31)} & \multicolumn{1}{c|}{1.46(0.13)} & 2.23(0.28) \\ \hline
C -- Mixed-HyLaD (k=1)  & \multicolumn{1}{c|}{\textbf{3.95(0.58)}} & \multicolumn{1}{c|}{\textbf{4.15(0.97)}} 
& \multicolumn{1}{c|}{\textbf{1.70(0.07)}} & \textbf{0.45(0.06)} & \multicolumn{1}{c|}{4.11(0.58)} & \multicolumn{1}{c|}{6.85(2.48)} & \multicolumn{1}{c|}{1.54(0.50)} & \textbf{0.44(0.06)} \\ \hline
B -- Meta-HyLaD (k=7)   & \multicolumn{1}{c|}{/} & \multicolumn{1}{c|}{/} 
& \multicolumn{1}{c|}{/} & / & \multicolumn{1}{c|}{\textbf{2.86(0.13)}} & \multicolumn{1}{c|}{\textbf{2.05(0.09)}} & \multicolumn{1}{c|}{\textbf{0.28(0.00)}} & 0.45(0.05) \\ \hline
C -- Mixed-HyLaD (k=7)   & \multicolumn{1}{c|}{/} & \multicolumn{1}{c|}{/} 
& \multicolumn{1}{c|}{/} & / & \multicolumn{1}{c|}{\textbf{2.62(0.11)}} & \multicolumn{1}{c|}{\textbf{1.93(0.12)}} & \multicolumn{1}{c|}{\textbf{0.32(0.07)}} & \textbf{0.39(0.12)} \\ \hline
\end{tabular}}
\end{table*}

\subsection{Data \& Experimental Settings }
\label{subsec:exp:data}
We consider common benchmarks 
including three simple physics systems of Pendulum \cite{botev2021priors}, Mass Spring \cite{KVAE}, and Bouncing Ball (under gravity) \cite{botev2021priors}, 
and two more complex physics systems of Double Pendulum \cite{botev2021priors} and Rigid Rody \cite{gradsim}. 
To demonstrate feasibility towards more complex systems, we further consider 
an additional dataset of dynamic PET \cite{PET}. 
For each of the five physics system, we randomly sample the initial states and parameters of the governing dynamic function, and generate time-series of system states and the corresponding image observations by physical rendering in \cite{gradsim}. We refer to the dynamic functions used for the generation of data as \textit{full physics} functions,
and design \textit{partial physics} functions to represent our imperfect prior knowledge about the observed data reflecting a variety of potentially additive and multiplicative errors
as summarized in Table \ref{table:systems}: 
in Pendulum and Mass Spring, \textit{partial physics} lacks the knowledge about damping (additive error), in Bouncing Ball and Rigid Body, \textit{partial physics} only considers gravity/force in one direction/plane (multiplicative error); in Double Pendulum, \textit{partial physics} ignores the impact of the mass difference between two pendulums (multiplicative error).

To introduce heterogeneity in the underlying dynamics, in data generation we vary the parameters of the full physics equations in the components both present (blue) and absent (red) in the prior physics, 
with their training and test distributions and number of samples summarized in Table \ref{table:systems}. 
More details on each dataset are in Appendix \ref{appendix:data}.
We leave descriptions of the dynamic PET dataset in Section \ref{subsec:exp:pet}. 






\subsection{Identifiability of Physics. \textit{vs.} Neural Dynamics}
\label{subsec:exp:id}
We first evaluate the reconstruction \textit{vs}.\ learn-to-identify objectives for identifying physics-based \textit{vs.} neural latent dynamic functions,  
on the Pendulum dataset (see Table \ref{table:systems}).

\textbf{Models:} Here we consider the three identification schemes for learning hybrid latent dynamic functions as described in Section \ref{subsec:ID} (Fig.~\ref{fig:models}).
All three learning strategies share the same backbone architectures, with their architectural and implementation details provided in \cref{app:implmentation}. 
We consider $k=1$ and $k=7$ for learn-to-identify strategies, to assess potential impact associated with the 
number of samples used to identify the dynamic functions. 


\begin{figure*}[t]
     \centering
     \includegraphics[width=.9\linewidth]{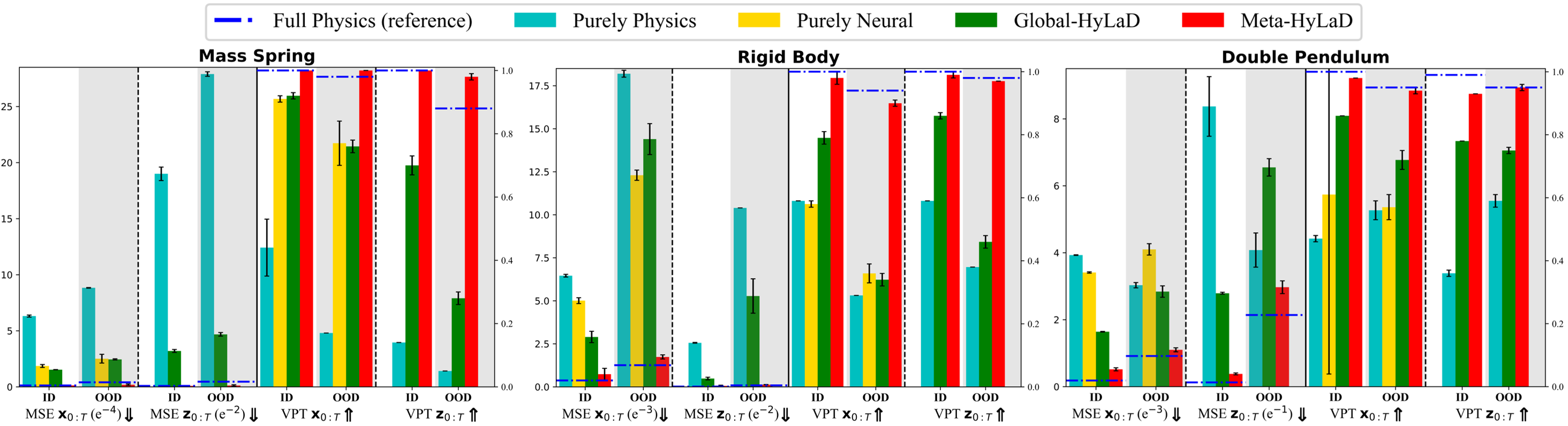}
     \caption{Comparison of alternative strategies for modeling the latent dynamic functions (purely physics, purely neural, Global-HyLaD, and Meta-HyLaD). MSE metrics are the lower the better, and VPT metrics are the higher the better.} 
     \label{fig:result2}
 \end{figure*}

\textbf{Metrics:} 
A successfully-identified latent dynamic function should be able to forecast a time series that is different from those
used to identify it
(but shares the same dynamics). 
To this end, 
we test the three learning strategies 
for two tasks:
1) a \textit{reconstruction} task to 
identify $\mathbf{c}_p$ and $\mathbf{c}_n$ from a test time-series
and reconstruct the seires itself, 
and 
2) a \textit{prediction} task 
where $\mathbf{c}_p$ and $\mathbf{c}_n$ 
are identified from a test time-series 
but used to predict another time-series
that follows the same dynamics but different initial conditions.

For both tasks, we measured the mean squared error (MSE) between the ground-truth and estimated time series at both the observation $\mathbf{x}_{0:T}$ and the state variable $\mathbf{z}_{0:T}$ level .
In addition, 
we consider two metrics
to separately evaluate how well $\mathbf{c}_p$ and $\mathbf{c}_n$ 
are each identified: 
for $\mathbf{c}_p$,
we directly measure the MSE between its identified and ground-truth value; 
for $\mathbf{c}_n$, 
because we do not expect its value to be physically meaningful, 
we instead measure the fit between the identified neural dynamic function $f_{\textrm{NN}_\theta}(\mathbf{z}_t;\mathbf{c}_n)$ 
and the ground truth residual function $-\beta \dot{\varphi}$ 
by their MSE. 

\textbf{Results: } 
As summarized in Table \ref{Experiment1_table}, 
all three learning strategies results in strong performance to \textit{reconstruct} the same time series used to identify the latent dynamic functions. 
However, 
latent dynamic functions identified with the reconstruction objective
(A) fail when 
attempting to predict different time series 
sharing the same dynamics. 
A deeper dive reveals something interesting: 
the physics component ($\mathbf{c}_p$) is well identified 
by all identification strategies (A-C);
in contrast, 
$\mathbf{c}_n$ is only successfully identified 
by the learn-to-identify strategy (B-C). 
Moreover, 
the mixed identification  strategy (C) seems to facilitate a better identification of $\mathbf{c}_n$ when $k=1$, 
although both 
Meta- (B) and Mixed-HyLaD (C) achieve similar performance when $k$ is increased to 7.
Interestingly, increasing $k$ for identifying $\mathbf{c}_n$ also improves
the identification of $\mathbf{c}_p$ in Mixed-HyLaD (C). 
We provide visual examples of identified 
$f_{\textrm{NN}_\phi}$ in \cref{appendix: result-sec4.2}. 



These experiments provide two important insights: 
1) a good reconstruction is not sufficient evidence that the underlying latent dynamics is correctly identified, and 2) the inclusion of physics inductive bias or the learn-to-identify strategy are two effective solutions to address the issue of \textit{identifiability}.





\subsection{Benefits of Hybrid Dynamic Functions}
\label{subsec:exp:hybrid}

We now evaluate the benefits of hybrid latent dynamic functions 
on the five physics systems described in Section \ref{subsec:exp:data}.


\textbf{Models:} 
We now fix the learn-to-identify strategy 
to that of Fig.~\ref{fig:models}B,
and 
compare four alternatives of latent dynamic functions: 
using only the \textit{partial physics} function (purely physics), 
using only a neural network (purely neural), 
hybrid as defined in Equation (\ref{eqn:hybrid}) but 
with a global $f_{\textrm{NN}_\phi}$ (Global-HyLaD), 
and Meta-HyLaD. 
We also obtain results on a latent dynamic function utilizing the \textit{full physics}, 
setting a performance reference 
when prior knowledge is perfect. 

\textbf{Metrics:} 
We consider 
MSE for both the observed images $\mathbf{x}_{0:T}$ and state variables $\mathbf{z}_{0:T}$, 
and 
Valid Prediction Time (VPT) that measures how long the predicted object’s trajectory remains close to the ground truth trajectory based on the MSE \cite{botev2021priors}. 
We separately evaluate these metrics 
for test time-series with parameters within (in distribution / ID) or outside (out of distribution / OOD) those used in the training data, as summarized in Table \ref{table:systems}.


 \begin{figure*}[t]
     \centering
     \includegraphics[width=.9\linewidth]{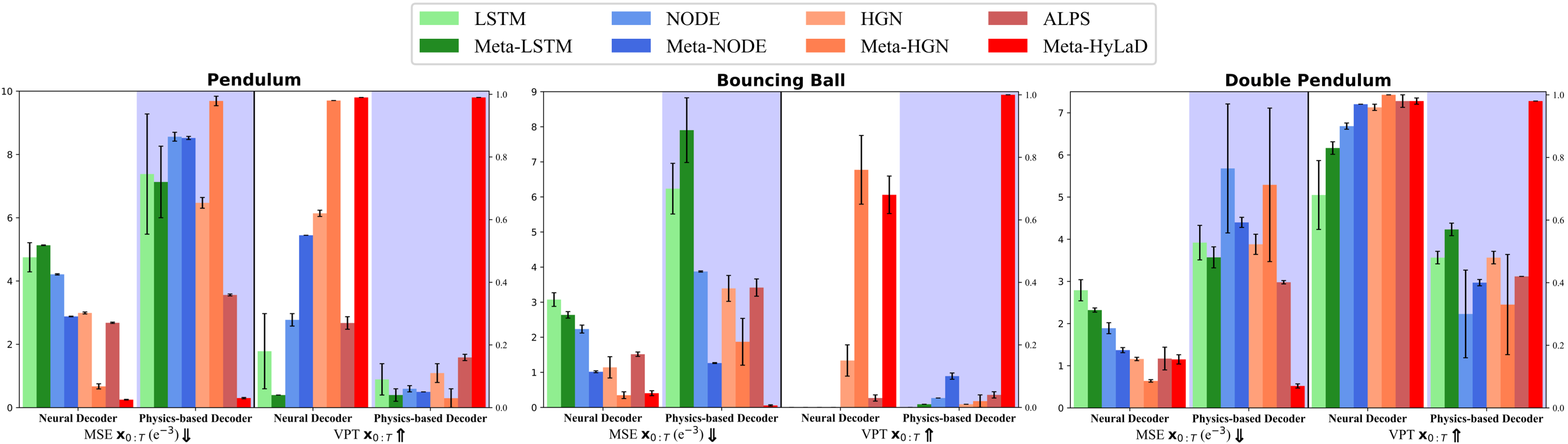}
     \caption{Comparison of Meta-HyLAD and baselines with neural and physics-based (blue-shaded background) decoders.}
     \label{fig:result3}
 \end{figure*}

\textbf{Results:} 
Fig.~\ref{fig:result2} summarizes the results on three datasets, 
with complete results 
and visual examples provided in \cref{appendix:4.3}: note that the purely neural dynamic functions (yellow), due to the abstract nature of its latent state variables, results in poor MSE/VPT metrics on $\mathbf{z}_{0:T}$ at a a magnitude larger than alternative approaches; their values are thus omitted 
from Fig.~\ref{fig:result2} but can be found in \cref{appendix:4.3}. 
As shown, 
purely physics-based approaches, when given perfect knowledge, achieve excellent performance as a reference (dashed blue line). 
Their accuracy, however, deteriorates substantially given imperfect knowledge (aqua), often to a level on par with purely neural approaches (yellow). 
The use of hybrid functions helps, 
although to a limited extent if the neural components are globally optimized (green).
Meta-HyLaD (red), 
with an ability to identify the neural components from data, 
substantially improves over all three alternatives -- its performance is on par with, and sometimes better than, the performance achieved with perfect physics. 



\subsection{Comparison with Existing Baselines}
\label{subsec:exp:baselines}

On the five physics systems, we then compare Meta-HyLaD with existing unsupervised latent dynamic models.
Note that there is no existing hybrid latent dynamic models. 

\textbf{Models:} We consider
1) ALPS \cite{ALPS} which 
considers a physics-based 
latent dynamic function (here the \textit{partial physics})
and a reconstruction-based learning objective, 
and 2) three neural latent dynamic models ranging from those with minimal physics prior (LSTM and neural ODE / NODE) and strong prior (Hamiltonian generative network / HGN) \cite{botev2021priors}. 
We consider their original formulations \cite{botev2021priors} that optimize global latent dynamic functions 
as illustrated in Fig.~\ref{fig:overview}B, 
as well as their extension to the presented meta-learning framework representative of recent works in meta-learning latent dynamics \cite{jiang2022sequential,CoDA}. 
Architecture details of all baselines are provided in \cref{app:implmentation}. 
This provides a comprehensive coverage of prior arts in 
their choices of latent dynamic functions and identifications strateiges. 

We further note that different decoders were used in these original baselines: the neural latent dynamic models utilize a neural network as the decoder (neural decoder) \cite{botev2021priors}; 
the latent-HGN utilizes a neural decoder but specifically uses the \textit{position} latent state as the input (neural decoder with prior) \cite{botev2021priors}; 
ALPS uses both this and a physics-based decoder \cite{ALPS}. 
To isolate the effect of decoders, 
we further evaluate each of these baselines using each of the three different decoders.

 
\textbf{Metrics:} 
We consider 
MSE and VPT for $\mathbf{x}_{0:T}$, 
and omit metrics on $\mathbf{z}_{0:T}$ due to its abstract nature in 
most baselines. 

 \begin{figure*}[t]
     \centering
     \includegraphics[width=.9\linewidth]{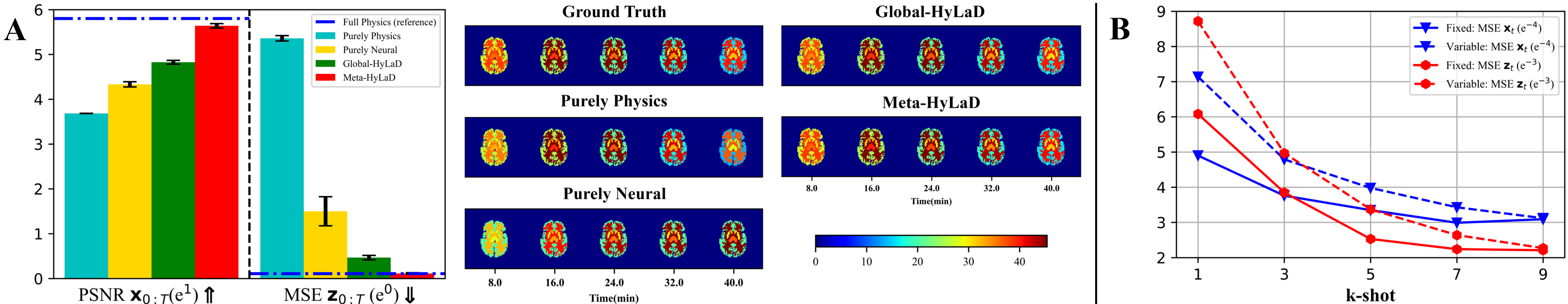}
     \caption{A: Quantitative metrics and examples in dynamic PET.
     B: Ablation with context-size $k$.}
     \label{fig:result4}
 \end{figure*}

\textbf{Results:} 
Fig.~\ref{fig:result3} summarizes the results on three datasets for physics-based \textit{vs}.\ neural decoders, 
with the complete quantitative results 
in \cref{appendix: result-sec4.4}.
Notably, across all datasets, Meta-HyLaD in general demonstrates a significant margin of improvements over all baselines including those utilizing meta-learning. 
This improvement is the most significant when used in combination of a physics-based decoder while, 
with a neural decoder with or without prior, 
Meta-HyLaD sometimes becomes comparable with the meta-extension of HGN
-- this again demonstrates the benefits of physics inductive bias in both models. Note that, as we will show in Section \ref{subsec:exp:pet}, 
Meta-HyLaD is more generally applicable
beyond Hamiltonian systems for which HGN is designed.

Interestingly, 
Meta-HyLaD improves with using a physics-based decoder, while the other baselines deteriorates. 
The effect of using a prior with the decoder (\cref{appendix:4.3}) is less consistent and varies with the dataset for all models.



\subsection{Feasibility on Biomedical Systems: Dynamic PET}
\label{subsec:exp:pet}

Finally, 
we consider 
recovering the spatial distribution and kinetics of radiotracer-labeled biological substrates from dynamic PET images.  
The regional tracker kinetics underlying PET 
can be described by the widely used compartment models
in Table \ref{table:systems}.
The total concentration of radioactivity in all tissues, $C_{T_i}(t)$, on the scanning time 
interval $[t_{k-1}, t_k]$ is measured as: $x_{ik}(t)=\frac{1}{t_k-t_{k-1}}\int_{t_{k-1}}^{t_k}C_{Ti}(t)e^{-\frac{t}{\tau}}dt$ at each voxel $i$.
The activity image $\mathbf{x}_k$ is obtained by lexicographic ordering of 
$x_{ik}$
at different voxels. 
While raw PET data are sinograms, in this proof-of-concept we consider 
$\mathbf{x}_{0:T}$ as the observed image series, with the goal to identify the compartment models. More details are in \cref{appendix:data-PET}.


\textbf{Data:}
We use the two-tissue compartment model listed in Table \ref{table:systems} to generate $\textbf{x}_{0:T}$ 
of a brain with 5 regions of interesting for 0-40 minutes with a scanning period of 0.5 minutes,
resulting in 80 frames per time series. 
2000 time-series samples are generated with  training/test and ID/OOD split of parameter ranges 
provided in \cref{appendix:data-PET}.

We then assume the one-tissue compartment model to be our prior physics in Meta-HyLaD. 
Compared to the additive and multiplicative errors considered earlier, this setup provides a more challenging scenario where the prior physics represents a crude approximation, 
\textit{i.e.,} $C_{Ti}(t)=C_{Ei}(t)+C_{Mi}(t)$ , for the data-generating latent dynamics.


\textbf{Model \& Metrics:} We compare Meta-HyLaD with purely physics, purely neural, 
and Global-HyLaD similar to  Section \ref{subsec:exp:hybrid}. 
We consider MSE and peak signal-to-noise-ratio (PSNR) on 
$\mathbf{x}_{0:T}$, and MSE and VPT on $\mathbf{z}_{0:T}$, \textit{i.e.}, \small{$C_{T}(0:T)$}. \normalsize

\textbf{Results:} 
Fig.~\ref{fig:result4}A summarize the quantitative metrics along with visual examples, with the full results and examples of tracer kinetics in \cref{appendix:data-PET} . 
The same observations in the physics systems still hold: Meta-HyLaD is the only model that is able to approach the reference performance (where data-generating physics is used in the identification), 
significantly outperforming the alternatives. This leaves exciting future real-world opportunities for Meta-HyLaD.

\section{Conclusions \& Discussion}
We present Meta-HyLaD as a first solution to unsupervised learning of hybrid latent dynamics, demonstrating the use of physics inductive bias combined with the learning-to-identify strategy to effectively leverage prior knowledge while identifying its unknown gap to observed data. 

\textbf{The effect of the size of context set:} Fig.~\ref{fig:result4}B shows that, 
while increasing the context-set size increases the performance of Meta-HyLaD, at $k=1$ it still significantly outperforms the various baselines. 
It is also minimally affected when trained with variable size of $k$, enhancing its ability to accommodate varying availability of context samples at test time.
If there is no knowledge about which samples share the same dynamics, 
future solutions may be to replace the averaging function in context-set embedding with 
attention mechanisms, to learn to extract similar context samples.   
Additional ablation results are provided in 
\cref{appendix: result-sec4.6}.


\textbf{Generality and failure modes:} 
In \cref{app:subsec:generality}, we provide examples that Meta-HyLaD can accommodate general design choices of $f_{\textrm{NN}_\phi}$ and 
maintains its strong identification and forecasting performance.
We also probe the potential failure modes for Meta-HyLaD and its use of decoders. 
As shown in \cref{app:subsec:failure_hylad},
if $f_{p}$ representing prior physics becomes too weak, 
\textit{e.g.,} modeling only the dampening effect on Pendulum, 
Meta-HyLaD may degenerate to a performance similar to purely neural approaches for forecasting $\mathbf{x}_t$, although still with a substantial gain in the accuracy of $\mathbf{z}_t$. 
Finally, 
while neural decoders provide competitive performance for all models considered, 
they could fail if the data-generating decoder function is not global.
In this setting,  
we show that the same learn-to-identify strategy can be extended to adapt the decoder (\cref{app:subsec:failure_decoder}), leaving another future avenue of Meta-HyLaD to be pursued.

\section*{Acknowledgements}
This work is supported in part by the National Key Research and Development Program of China(No: 2020AAA0109502); the Talent Program of Zhejiang Province (No: 2021R51004); NIH NHLBI grant R01HL145590 and NSF OAC-2212548. This paper presents work whose goal is to advance the field of Machine Learning. There are many potential societal consequences of our work, none which we feel must be specifically highlighted here.



\nocite{Nocite1, Nocite2, Nocite3, Nocite4, Nocite5, Nocite6, Nocite7}
\normalem
\bibliography{reference}
\bibliographystyle{splncs04}

\newpage
\appendix
\onecolumn
\section{Data Details and Experimental Settings of Physics Systems}
\label{appendix:data}
\begin{figure}[h]
    \centering
    \includegraphics[width=\linewidth]{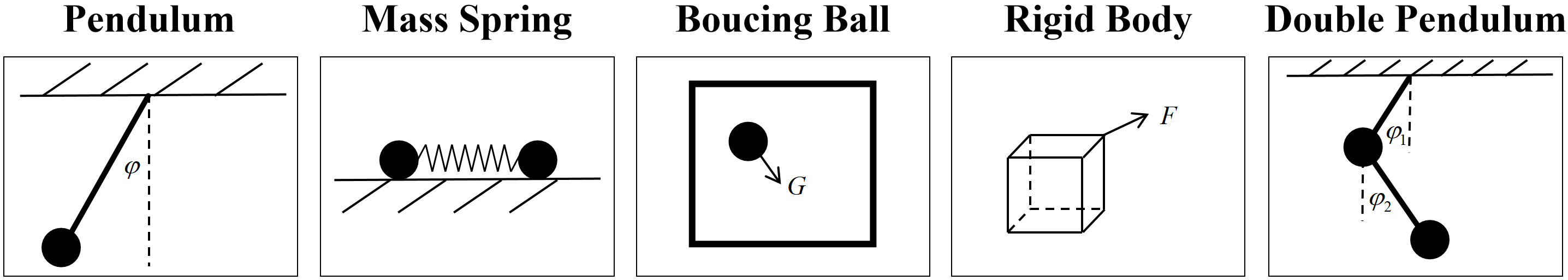}
    \caption{The overview of five physics systems}
    \label{Physics Systems}
\end{figure}

Here we provide complete data details and experimental settings of physics systems and the overview of physics systems can be seen in Fig. \ref{Physics Systems}. The ratio of training samples:ID testing samples:OOD testing samples is roughly 6:2:2 in all datasets.
\subsection{Data Details}
\label{appendix:data-physics}
\textbf{(1) Pendulum: }The dynamic equation is: $\frac{\mathrm{d}}{\mathrm{d} t}\begin{bmatrix} \varphi  \\ \dot{\varphi}  \end{bmatrix}= \begin{bmatrix} \dot{\varphi}  \\ -\frac{{\color{blue}{G}}}{L}\sin(\varphi)-{{\color{red}{\beta}}} \dot{\varphi} \end{bmatrix}$, where $\varphi$ is the angle, $\dot{\varphi}$ is the angular velocity, ${\color{blue}{G}}, {\color{red}{\beta}}$ are the gravitational constant and the damping coefficient, and $L$ is the length of the pendulum. We fix $L=2.0$, then sample $\varphi_0 \in[-\frac{1}{2}\pi, \frac{1}{2}\pi]$, $\dot{\varphi}_0 \in [-2.0, 2.0]$, ${\color{blue}{G}} \in [5.0, 15.0]$, ID:${\color{red}{\beta}}\in(0.0, 1.0]$ and OOD: ${\color{red}{\beta}}\in(1.0, 1.25]$. We generate a 25 step in time domain [0.0s, 5.0s] following the dynamic equation, and render corresponding 32 by 32 by 1 pixel observation snapshots. In total, we generate 15390 training and test sequences. \\
\textbf{(2) Mass Spring: }The dynamic equation is: $\frac{\mathrm{d}}{\mathrm{d} t}\begin{bmatrix} v_1  \\ v_2  \end{bmatrix}=
\begin{bmatrix} -\frac{\Tilde{x}}{\left | \Tilde{x} \right |} \frac{{\color{blue}{k}}}{m_1}
\left ( \left | \Tilde{x} \right | -l_0 \right ) - \frac{{\color{red}{\beta}}}{m_1} \Tilde{v} \\ \frac{\Tilde{x}}{\left | \Tilde{x} \right |} \frac{{\color{blue}{k}}}{m_2}
\left ( \left | \Tilde{x} \right | -l_0 \right ) +\frac{{\color{red}{\beta}}}{m_2} \Tilde{v}
\end{bmatrix}$, where $\Tilde{x}=x_1-x_2$, $\Tilde{v}=v_1-v_2$, $x_1, x_2$ are the positions of node$_1$ and node$_2$, $v_1, v_2$ are the velocities of node$_1$ and node$_2$ and ${\color{blue}{k}}, {\color{red}{\beta}} $ are the stiffness and damping coefficient of the spring and damper. We fix $l_0=6.0, m_1=m_2=1.0, v_{10}=v_{20}=0$, then sample $x_{10} \in [-4.0, -2.0], x_{2.0} \in [2.0, 4.0], {\color{blue}{k}} \in [5.0, 15.0]$, ID: ${\color{red}{\beta}}\in(0.0, 1.0]$ and OOD: ${\color{red}{\beta}}\in(1.0, 1.5]$. We generate a 25 step in time domain [0.0s, 5.0s] following the dynamic equation, and render corresponding 32 by 32 by 3 pixel observation snapshots. In total, we generate 12960 training and test sequences.
\\
\textbf{(3) Bouncing Ball: }The dynamic equation is: $\frac{\mathrm{d}}{\mathrm{d} t}\begin{bmatrix} v_x  \\ v_y  \end{bmatrix}=\begin{bmatrix}{{\color{blue}{G}}} \cos {\color{red}{\varphi}} \\{{\color{blue}{G}}}\sin {\color{red}{\varphi}}  \end{bmatrix}$, where ${\color{blue}{G}}, {\color{red}{\varphi}}$ are the amplitude and direction of gravity. When there is a collision, $v_x=-v_x$ or $v_y=-v_y$. We set the collision position at $x = 5.0, x = -5.0, y = 5.0, y = -5.0$. We fix $v_{x0}=v_{y0}=0$, then sample $x_0\in[-4.0, 4.0], y_0\in[-4.0, 4.0], {\color{blue}{G}}\in(0.0, 5.0]$, ${\color{red}{\varphi}}=\frac{{\color{red}{i}}}{8}\pi$, ID: ${\color{red}{i}}=[-6, -5, -4, -3, -2, -1, 2, 3, 4, 5, 6, 7]$ and OOD: ${\color{red}{i}}=[-8, -7, 0, 1]$. We generate a 25 step in time domain [0.0s, 2.5s] following the dynamic equation, and render corresponding 32 by 32 by 1 pixel observation snapshots. In total, we generate 21060 training and test sequences.
\\
\textbf{(4) Rigid Body: } We represent the state of a 3D cube as $\mathbf{q}=
\left [ \mathbf{x},\mathbf{r} \right ]$ consisting of a position $\mathbf{x}\in \mathbb{R}^3$ and a quaternion $\mathbf{r} \in \mathbb{R}^4$. The generalized velocity of the cube is $\mathbf{u}=[\mathbf{v}, \mathbf{\omega}]$ and its dynamic equation is: $\begin{bmatrix} m & 0\\ 0 & \mathbf{I} \end{bmatrix}
\begin{bmatrix}\dot{\mathbf{v}}\\ \dot{\mathbf{\omega}} \end{bmatrix} = 
\begin{bmatrix}\mathbf{f}\\  \mathbf{r}\times\mathbf{f} \end{bmatrix} -
\begin{bmatrix}0\\ \mathbf{\omega}\times \mathbf{I} \mathbf{\omega}\end{bmatrix}$, where $m, \mathbf{I}$ are the mass and inertia matrix, $\mathbf{f} =f\cdot [\cos{\color{blue}{\psi}}\sin{\color{red}{\varphi}} , \sin{\color{blue}{\psi}}\sin{\color{red}{\varphi}}, \cos{\color{red}{\varphi}}]$ and $f, {\color{blue}{\psi}}, {\color{red}{\varphi}}$ are the magnitude of the force, its direction in the xy-plane and its direction with respect to the z-axis. We fix $z_0=0, \mathbf{r}=[0, 0, 0, 1], \mathbf{v}=[0, 0, 0], \mathbf{\omega}=[0, 0, 0]$ $m=8$ and $f=10$, then sample $x_0\in[-1.0, 1.0], y_0\in[-1.0, 1.0], {\color{blue}{\psi}}\in[-\pi, \pi)$, ID: ${\color{red}{\varphi}}\in(\frac{1}{8}\pi, \frac{1}{2}\pi]$ and OOD: ${\color{red}{\varphi}}\in[0, \frac{1}{8}\pi]$. We generate a 30 step in time domain [0.0s, 1.0s] following the dynamic equation, and render corresponding 64 by 64 by 3 pixel observation snapshots. In total, we generate 4800 training and test sequences.
\\
\textbf{(5) Double Pendulum: } The dynamic equation is: $\frac{d}{dt}\begin{bmatrix} \dot{\varphi}_1  \\ \dot{\varphi}_2 \end{bmatrix}=
\begin{bmatrix}\frac{{\color{blue}{G}}\sin\varphi_2\phi_1 -\phi_2(L_1\dot{\varphi}_1^2\phi_1+L_2\dot{\varphi}_2^2)
-({\color{red}{\Tilde{m}}}+1){\color{blue}{G}}\sin\varphi_1}{L_1({\color{red}{\Tilde{m}}}+\phi_2^2)} \\
\frac{({\color{red}{\Tilde{m}}}+1)(L_1\dot{\varphi}_1^2\phi_2 -{\color{blue}{G}}\sin\varphi_2+{\color{blue}{G}}\sin\varphi_1\phi_1)
+L_2\dot{\varphi}_2^2\phi_1 \phi_2}{L_2({\color{red}{\Tilde{m}}}+\phi_2^2)}\end{bmatrix}$, where ${\color{red}{\Tilde{m}}}=m_1/m_2$,
$\phi_1=\cos(\varphi_1 - \varphi_2)$, $\phi_2=\sin(\varphi_1 - \varphi_2)$ and $m_1, m_2$ are the mass of two pendulums. We fix $L_1=L_2=1.5, \dot{\varphi}_{10}=\dot{\varphi}_{10}=0$, then sample $\varphi_{10}\in[-\frac{1}{2}\pi, \frac{1}{2}\pi], \varphi_{20}\in[-\frac{1}{8}\pi, \frac{1}{8}\pi], {\color{blue}{G}}\in[5.0, 15.0]$, ID: ${\color{red}{\Tilde{m}}}\in[0.5, 1.5]$ and OOD: ${\color{red}{\Tilde{m}}}\in(1.5, 2.0]$. We generate a 40 step in time domain [0.0s, 4.0s] following the dynamic equation, and render corresponding 32 by 32 by 3 pixel observation snapshots. In total, we generate 12705 training and test sequences.
\subsection{Experimental Settings for \cref{subsec:exp:hybrid}}
Here we provide specific forms of different latent dynamic functions in \cref{subsec:exp:hybrid} for each datasets. 
\subsubsection{Pendulum}
\begin{table}[h]
\caption{Experimental Setting of Pendulum}
\centering
\resizebox{0.48\linewidth}{!}{
\begin{tabular}{ll}
\hline
\multicolumn{1}{c}{Dynamics} & \multicolumn{1}{c}{Equation} \\ \hline
Full Physics  &  $\frac{\mathrm{d}}{\mathrm{d} t}\begin{bmatrix} \varphi  \\ \dot{\varphi}  \end{bmatrix}= \begin{bmatrix} \dot{\varphi}  \\ -\frac{{\color{blue}{G}}}{L}\sin(\varphi)- {\color{red}{\beta}}  \dot{\varphi} \end{bmatrix}$ \\
Purely Physics &  $\frac{\mathrm{d}}{\mathrm{d} t}\begin{bmatrix} \varphi  \\ \dot{\varphi}  \end{bmatrix}= \begin{bmatrix} \dot{\varphi}  \\ -\frac{{\color{blue}{G}}}{L}\sin(\varphi)\end{bmatrix}$ \\
Purely Neural &  $\frac{\mathrm{d}}{\mathrm{d} t}\begin{bmatrix} \varphi  \\ \dot{\varphi}  \end{bmatrix}= f_{\text{NN}_\phi}(\varphi, \dot{\varphi}; \mathbf{c}_n)$ \\
Global-HyLaD       &  $\frac{\mathrm{d}}{\mathrm{d} t}\begin{bmatrix} \varphi  \\ \dot{\varphi}  \end{bmatrix}= \begin{bmatrix} \dot{\varphi}  \\ -\frac{{\color{blue}{G}}}{L}\sin(\varphi) \end{bmatrix}+f_{\text{NN}_\phi}(\dot{\varphi})$ \\ 
Meta-HyLaD         &  $\frac{\mathrm{d}}{\mathrm{d} t}\begin{bmatrix} \varphi  \\ \dot{\varphi}  \end{bmatrix}= \begin{bmatrix} \dot{\varphi}  \\ -\frac{{\color{blue}{G}}}{L}\sin(\varphi) \end{bmatrix}+f_{\text{NN}_\phi}(\dot{\varphi}; \mathbf{c}_n)$ \\
\hline
\end{tabular}}
\end{table}
\subsubsection{Mass Spring}
\begin{table}[h]
\centering
\caption{Experimental Setting of Mass Spring}
\resizebox{0.55\linewidth}{!}{
\begin{tabular}{ll}
\hline
\multicolumn{1}{c}{Dynamics} & \multicolumn{1}{c}{Equation} \\ \hline
Full Physics & $\frac{\mathrm{d}}{\mathrm{d} t}\begin{bmatrix} v_1  \\ v_2  \end{bmatrix}=
\begin{bmatrix} -\frac{\Tilde{x}}{\left | \Tilde{x} \right |} \frac{{\color{blue}{k}}}{m_1}
\left ( \left | \Tilde{x} \right | -l_0 \right ) - \frac{{\color{red}{\beta}}}{m_1} \Tilde{v} \\ \frac{\Tilde{x}}{\left | \Tilde{x} \right |} \frac{{\color{blue}{k}}}{m_2}
\left ( \left | \Tilde{x} \right | -l_0 \right ) +\frac{{\color{red}{\beta}}}{m_2} \Tilde{v}
\end{bmatrix}$   
\\
Purely Physics & $\frac{\mathrm{d}}{\mathrm{d} t}\begin{bmatrix} v_1  \\ v_2  \end{bmatrix}=
\begin{bmatrix} -\frac{\Tilde{x}}{\left | \Tilde{x} \right |} \frac{{\color{blue}{k}}}{m_1}
\left ( \left | \Tilde{x} \right | -l_0 \right ) \\ \frac{\Tilde{x}}{\left | \Tilde{x} \right |} \frac{{\color{blue}{k}}}{m_2}
\left ( \left | \Tilde{x} \right | -l_0 \right )
\end{bmatrix}$  
\\
Purely Neural & $\frac{\mathrm{d}}{\mathrm{d} t}\begin{bmatrix} v_1  \\ v_2  \end{bmatrix}=
f_{\text{NN}_\phi}(x_1,x_2,v_1,v_2;\mathbf{c}_n)$
\\
Global-HyLaD  & $\frac{\mathrm{d}}{\mathrm{d} t}\begin{bmatrix} v_1  \\ v_2  \end{bmatrix}=
\begin{bmatrix} -\frac{\Tilde{x}}{\left | \Tilde{x} \right |} \frac{{\color{blue}{k}}}{m_1}
\left ( \left | \Tilde{x} \right | -l_0 \right ) \\ \frac{\Tilde{x}}{\left | \Tilde{x} \right |} \frac{{\color{blue}{k}}}{m_2}
\left ( \left | \Tilde{x} \right | -l_0 \right )
\end{bmatrix}+f_{\text{NN}_\phi}(v_1,v_2)$ 
\\ 
Meta-HyLaD & $\frac{\mathrm{d}}{\mathrm{d} t}\begin{bmatrix} v_1  \\ v_2  \end{bmatrix}=
\begin{bmatrix} -\frac{\Tilde{x}}{\left | \Tilde{x} \right |} \frac{{\color{blue}{k}}}{m_1}
\left ( \left | \Tilde{x} \right | -l_0 \right ) \\ \frac{\Tilde{x}}{\left | \Tilde{x} \right |} \frac{{\color{blue}{k}}}{m_2}
\left ( \left | \Tilde{x} \right | -l_0 \right )
\end{bmatrix}+f_{\text{NN}_\phi}(v_1,v_2;\mathbf{c}_n)$
\\ \hline
\end{tabular}}
\end{table}
\subsubsection{Bouncing Ball}
\begin{table}[h]
\centering
\caption{Experimental Setting of Bouncing Ball}
\resizebox{0.45\linewidth}{!}{
\begin{tabular}{ll}
\hline
\multicolumn{1}{c}{Dynamics} & \multicolumn{1}{c}{Equation} \\ \hline
Full Physics &  $\frac{\mathrm{d}}{\mathrm{d} t}\begin{bmatrix} v_x  \\ v_y  \end{bmatrix}=\begin{bmatrix}{{\color{blue}{G}}} \cos {\color{red}{\varphi}} \\{{\color{blue}{G}}}\sin {\color{red}{\varphi}}  \end{bmatrix}$
\\
Purely Physics & $\frac{\mathrm{d}}{\mathrm{d} t}\begin{bmatrix} v_x  \\ v_y  \end{bmatrix}=
\begin{bmatrix} 0  \\-{{\color{blue}{G}}}  \end{bmatrix}$
\\
Purely Neural & $\frac{\mathrm{d}}{\mathrm{d} t}\begin{bmatrix} v_x  \\ v_y  \end{bmatrix}= f_{\text{NN}_\phi}({{\color{blue}{G}}};\mathbf{c}_n)$
\\
Global-HyLaD  & $\frac{\mathrm{d}}{\mathrm{d} t}\begin{bmatrix} v_x  \\ v_y  \end{bmatrix}=
\begin{bmatrix} 0  \\-{{\color{blue}{G}}} \end{bmatrix}+f_{\text{NN}_\phi}({{\color{blue}{G}}})$
\\ 
Meta-HyLaD & $\frac{\mathrm{d}}{\mathrm{d} t}\begin{bmatrix} v_x  \\ v_y  \end{bmatrix}=
\begin{bmatrix} 0  \\-{{\color{blue}{G}}} \end{bmatrix}+f_{\text{NN}_\phi}({{\color{blue}{G}}};\mathbf{c}_n)$
\\ \hline
\end{tabular}}
\end{table}

\newpage
\subsubsection{Rigid Body}
\begin{table}[h]
\centering
\caption{Experimental Setting of Rigid Body}
\resizebox{0.5\linewidth}{!}{
\begin{tabular}{ll}
\hline
\multicolumn{1}{c}{Dynamics} & \multicolumn{1}{c}{Equation} \\ \hline
Full Physics &  $\mathbf{f} =f\cdot [\cos{\color{blue}{\psi}}\sin{\color{red}{\varphi}} , \sin{\color{blue}{\psi}}\sin{\color{red}{\varphi}}, \cos{\color{red}{\varphi}}]$
\\
Purely Physics & $\mathbf{f} =f\cdot [\cos{\color{blue}{\psi}} , \sin{\color{blue}{\psi}}, 0]$
\\
Purely Neural & 
$\mathbf{f}=f_{\text{NN}_\phi}({\color{blue}{\psi}};\mathbf{c}_n)$
\\
Global-HyLaD  & $\mathbf{f} =f\cdot [\cos{\color{blue}{\psi}} , \sin{\color{blue}{\psi}}, 0] + f_{\text{NN}_\phi}({\color{blue}{\psi}})$
\\ 
Meta-HyLaD & $\mathbf{f} =f\cdot [\cos{\color{blue}{\psi}} , \sin{\color{blue}{\psi}}, 0] + f_{\text{NN}_\phi}({\color{blue}{\psi}};\mathbf{c}_n)$
\\ \hline
\end{tabular}}
\end{table}
\subsubsection{Double Pendulum}
\begin{table}[h]
\centering
\caption{Experimental Setting of Double Pendulum}
\resizebox{0.75\linewidth}{!}{
\begin{tabular}{ll}
\hline
\multicolumn{1}{c}{Dynamics} & \multicolumn{1}{c}{Equation} \\ \hline
Full Physics & $\frac{d}{dt}\begin{bmatrix} \dot{\varphi}_1  \\ \dot{\varphi}_2 \end{bmatrix}=
\begin{bmatrix}\frac{{\color{blue}{G}}\sin\varphi_2\phi_1 -\phi_2(L_1\dot{\varphi}_1^2\phi_1+L_2\dot{\varphi}_2^2)
-({\color{red}{\Tilde{m}}}+1){\color{blue}{G}}\sin\varphi_1}{L_1({\color{red}{\Tilde{m}}}+\phi_2^2)} \\
\frac{({\color{red}{\Tilde{m}}}+1)(L_1\dot{\varphi}_1^2\phi_2 -{\color{blue}{G}}\sin\varphi_2+{\color{blue}{G}}\sin\varphi_1\phi_1)
+L_2\dot{\varphi}_2^2\phi_1 \phi_2}{L_2({\color{red}{\Tilde{m}}}+\phi_2^2)}\end{bmatrix}$
\\
Purely Physics & $\frac{d}{dt}\begin{bmatrix} \dot{\varphi}_1  \\ \dot{\varphi}_2 \end{bmatrix}=
\begin{bmatrix}\frac{{\color{blue}{G}}\sin\varphi_2\phi_1 -\phi_2(L_1\dot{\varphi}_1^2\phi_1+L_2\dot{\varphi}_2^2)
-2{\color{blue}{G}}\sin\varphi_1}{L_1(1+\phi_2^2)} \\
\frac{2(L_1\dot{\varphi}_1^2\phi_2 -{\color{blue}{G}}\sin\varphi_2+{\color{blue}{G}}\sin\varphi_1\phi_1)
+L_2\dot{\varphi}_2^2\phi_1 \phi_2}{L_2(1+\phi_2^2)}\end{bmatrix}$
\\
Purely Neural & 
$\frac{d}{dt}\begin{bmatrix} \dot{\varphi}_1  \\ \dot{\varphi}_2 \end{bmatrix}=f_{\text{NN}_\phi}(\varphi_1, \varphi_2, \dot{\varphi}_1, \dot{\varphi}_2;\mathbf{c}_n)$
\\
Global-HyLaD  & 
$\frac{d}{dt}\begin{bmatrix} \dot{\varphi}_1  \\ \dot{\varphi}_2 \end{bmatrix}=
\begin{bmatrix}\frac{{\color{blue}{G}}\sin\varphi_2\phi_1 -\phi_2(L_1\dot{\varphi}_1^2\phi_1+L_2\dot{\varphi}_2^2)
-2{\color{blue}{G}}\sin\varphi_1}{L_1(1+\phi_2^2)} \\
\frac{2(L_1\dot{\varphi}_1^2\phi_2 -{\color{blue}{G}}\sin\varphi_2+{\color{blue}{G}}\sin\varphi_1\phi_1)
+L_2\dot{\varphi}_2^2\phi_1 \phi_2}{L_2(1+\phi_2^2)}\end{bmatrix}+f_{\text{NN}_\phi}(\varphi_1, \varphi_2, \dot{\varphi}_1, \dot{\varphi}_2)$
\\ 
Meta-HyLaD & 
$\frac{d}{dt}\begin{bmatrix} \dot{\varphi}_1  \\ \dot{\varphi}_2 \end{bmatrix}=
\begin{bmatrix}\frac{{\color{blue}{G}}\sin\varphi_2\phi_1 -\phi_2(L_1\dot{\varphi}_1^2\phi_1+L_2\dot{\varphi}_2^2)
-2{\color{blue}{G}}\sin\varphi_1}{L_1(1+\phi_2^2)} \\
\frac{2(L_1\dot{\varphi}_1^2\phi_2 -{\color{blue}{G}}\sin\varphi_2+{\color{blue}{G}}\sin\varphi_1\phi_1)
+L_2\dot{\varphi}_2^2\phi_1 \phi_2}{L_2(1+\phi_2^2)}\end{bmatrix}+f_{\text{NN}_\phi}(\varphi_1, \varphi_2, \dot{\varphi}_1, \dot{\varphi}_2;\mathbf{c}_n)$
\\ \hline
\end{tabular}}
\end{table}

\section{Examples of Identified Neural Dynamic Functions (Additioanl Results for \cref{subsec:exp:id})}
\label{appendix: result-sec4.2}
\begin{figure}[h]
    \centering
    \includegraphics[width=.75\linewidth]{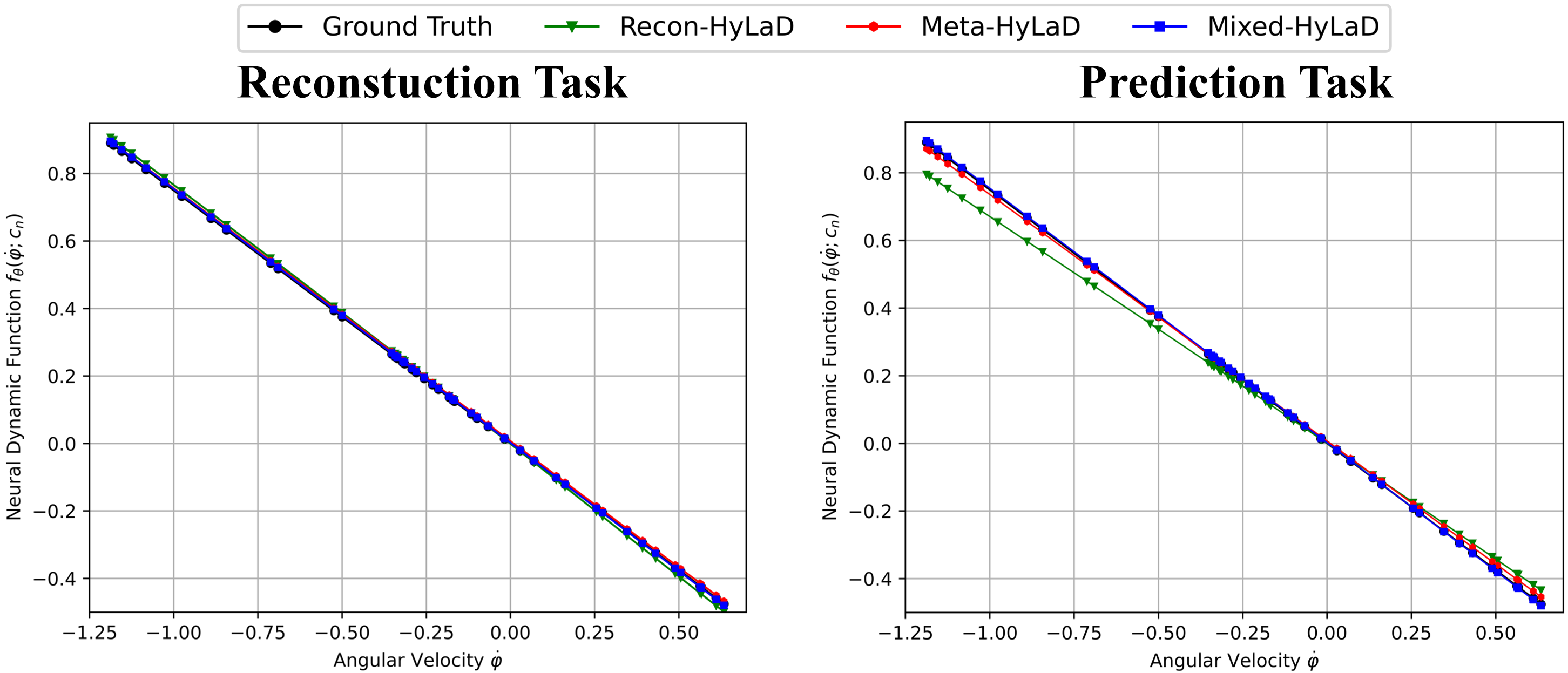}
    \caption{Visualization of different learning strategies}
    \label{Result:Appendix-4.2}
\end{figure}

We also visualize the neural dynamic function $f_{\text{NN}_\theta}(\mathbf{z}_t;\mathbf{c}_n)$ of three learning strategies and ground truth $-\beta\dot{\varphi}$, see Fig \ref{Result:Appendix-4.2}. 
From figure, we can clearly see that all three learning strategies perform well on reconstruction task, but Recon-HyLaD will fail on prediction task, which shows that a good reconstruction is not sufficient evidence that the underlying latent dynamics is correctly identified and the learn-to-identify strategy is an effective solution to address this issue of identifiability.

\newpage

\section{Additional Results of \cref{subsec:exp:hybrid}: Benefits of Hybrid Dynamics}
\label{appendix:4.3}

Here we provide complete experimental results and some visual examples of \cref{subsec:exp:hybrid}.
\subsection{Pendulum}
\begin{table}[h]
\centering
\caption{Results of Pendulum}
\begin{tabular}{llcccc}
\hline
\multicolumn{1}{c}{Dynamics} & \multicolumn{1}{c}{Data} & MSE $\mathbf{x}_t$ (e$^{-3}$)$\downarrow$  & MSE $\mathbf{z}_t$ (e$^{-2}$) $\downarrow$ & VPT $\mathbf{x}_t$ $\uparrow$ & VPT of $\mathbf{z}_t$ $\uparrow$ \\ \hline
\multirow{2}{*}{\begin{tabular}[c]{@{}l@{}}Full Physics\\ (oracle)\end{tabular}}
& ID & 0.31(0.05) & 0.24(0.06) & 0.99(0.00) & 0.98(0.01) \\
& OOD & 0.62(0.06) & 0.48(0.08) & 0.97(0.01) & 0.88(0.03) \\ 
\hline
\multirow{2}{*}{Purely Physics} 
& ID  & 4.38(0.00) & 13.09(0.18) & 0.13(0.19) & 0.05(0.01) \\ 
& OOD & 4.94(0.07) & 19.57(0.55) & 0.05(0.00) & 0.01(0.00) \\ \hline
\multirow{2}{*}{Purely Neural} 
& ID & 3.36(0.52) & 109.96(54.82) & 0.53(0.07) & 0.00(0.00) \\
& OOD & 4.02(0.54) & 78.47(35.18) & 0.43(0.03) & 0.00(0.00) \\ \hline
\multirow{2}{*}{Global-HyLaD} 
& ID & 2.26(0.05) & 4.01(0.05) & 0.60(0.01) & 0.41(0.01) \\ 
& OOD  & 2.99(0.06) & 6.50(0.02) & 0.38(0.02) & 0.17(0.00)  \\ \hline
\multirow{2}{*}{Meta-HyLaD} 
& ID & \textbf{0.30(0.01)} & \textbf{0.22(0.00)} & \textbf{0.99(0.00)} & \textbf{0.98(0.00)} \\ 
& OOD & \textbf{0.25(0.02)} & \textbf{0.17(0.02)} & \textbf{0.99(0.00)} & \textbf{0.95(0.01)} \\ \hline
\end{tabular}
\end{table}

\begin{figure}[h]
    \centering
    \includegraphics[width=0.95\linewidth]{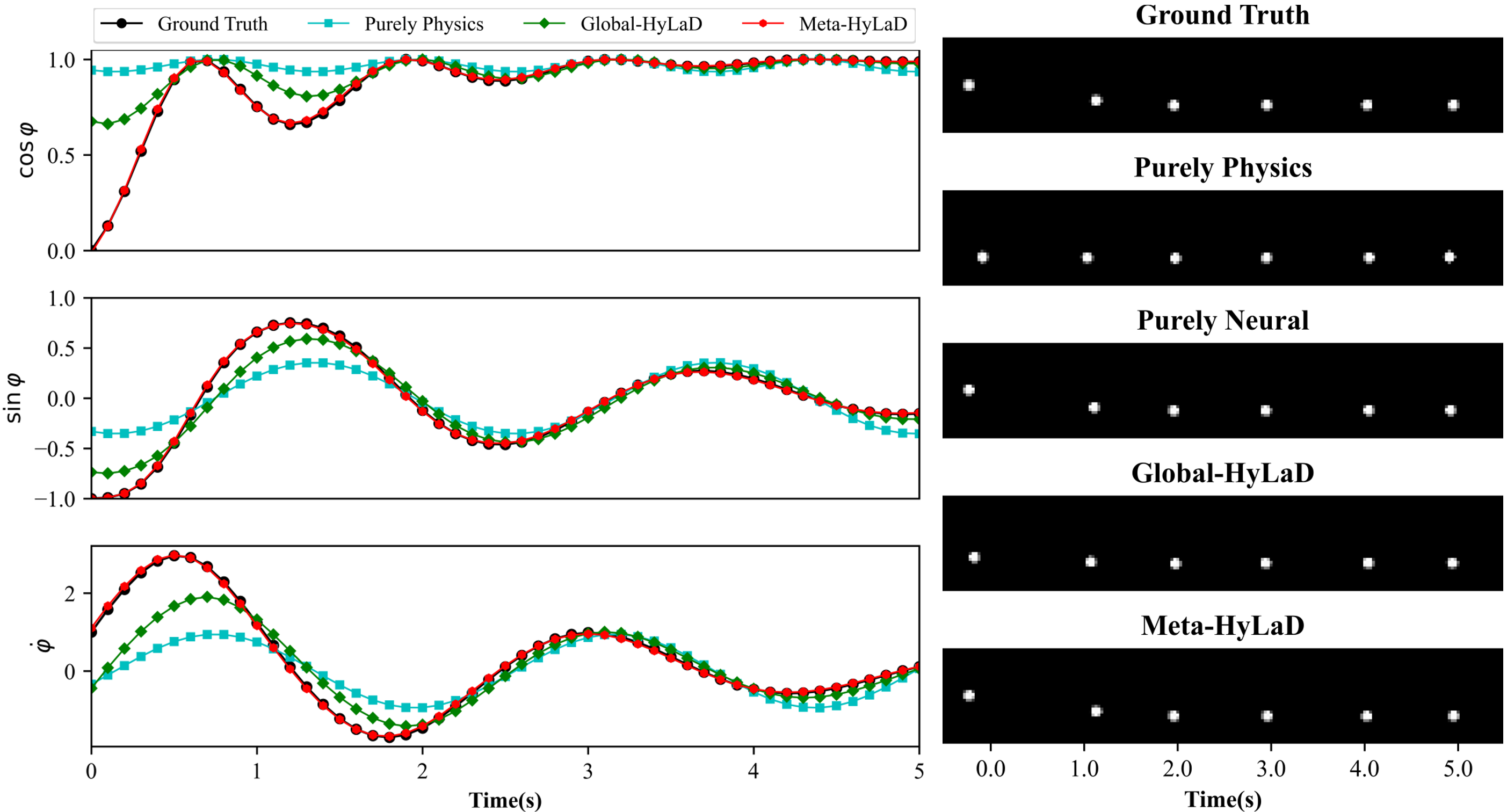}
    \caption{Visual examples of Pendulum}
\end{figure}

\newpage
\subsection{Mass Spring}
\begin{table}[h]
\centering
\caption{Results of Mass Spring}
\begin{tabular}{llcccc}
\hline
\multicolumn{1}{c}{Dynamics} & \multicolumn{1}{c}{Data} & MSE $\mathbf{x}_t$ (e$^{-4}$)$\downarrow$  & MSE $\mathbf{z}_t$ (e$^{-2}$) $\downarrow$ & VPT $\mathbf{x}_t$ $\uparrow$ & VPT of $\mathbf{z}_t$ $\uparrow$ \\ \hline
\multirow{2}{*}{\begin{tabular}[c]{@{}l@{}}Full Physics\\ (oracle)\end{tabular}}
& ID & 0.12(0.01) & 0.09(0.02)& 1.00(0.00) & 1.00(0.00) \\
& OOD & 0.40(0.03) & 0.46(0.07) & 0.98(0.01) & 0.88(0.02) \\ \hline
\multirow{2}{*}{Purely Physics} 
& ID & 6.32(0.09) & 19.01(0.56) & 0.44(0.09) & 0.14(0.00) \\ 
& OOD & 8.83(0.03) & 27.93(0.21) & 0.17(0.00) & 0.05(0.00) \\ \hline
\multirow{2}{*}{Purely Neural} 
& ID & 1.86(0.13) & 393.47(75.06) & 0.91(0.01) & 0.05(0.00) \\
& OOD & 2.51(0.39) & 348.02(93.33) & 0.77(0.07) & 0.05(0.00) \\ \hline
\multirow{2}{*}{Global-HyLaD} 
& ID & 1.53(0.01) & 3.20(0.13) & 0.92(0.01) & 0.70(0.03) \\ 
& OOD  & 2.45(0.05) & 4.68(0.16) & 0.76(0.02) & 0.28(0.02)  \\ \hline
\multirow{2}{*}{Meta-HyLaD} 
& ID & \textbf{0.13(0.03)} & \textbf{0.09(0.02)} & \textbf{1.00(0.00)} & \textbf{1.00(0.00)} \\ 
& OOD & \textbf{0.19(0.07)} & \textbf{0.12(0.06)} & \textbf{1.00(0.00)} & \textbf{0.98(0.01)} \\ \hline
\end{tabular}
\end{table}

\begin{figure}[h]
    \centering
    \includegraphics[width=0.95\linewidth]{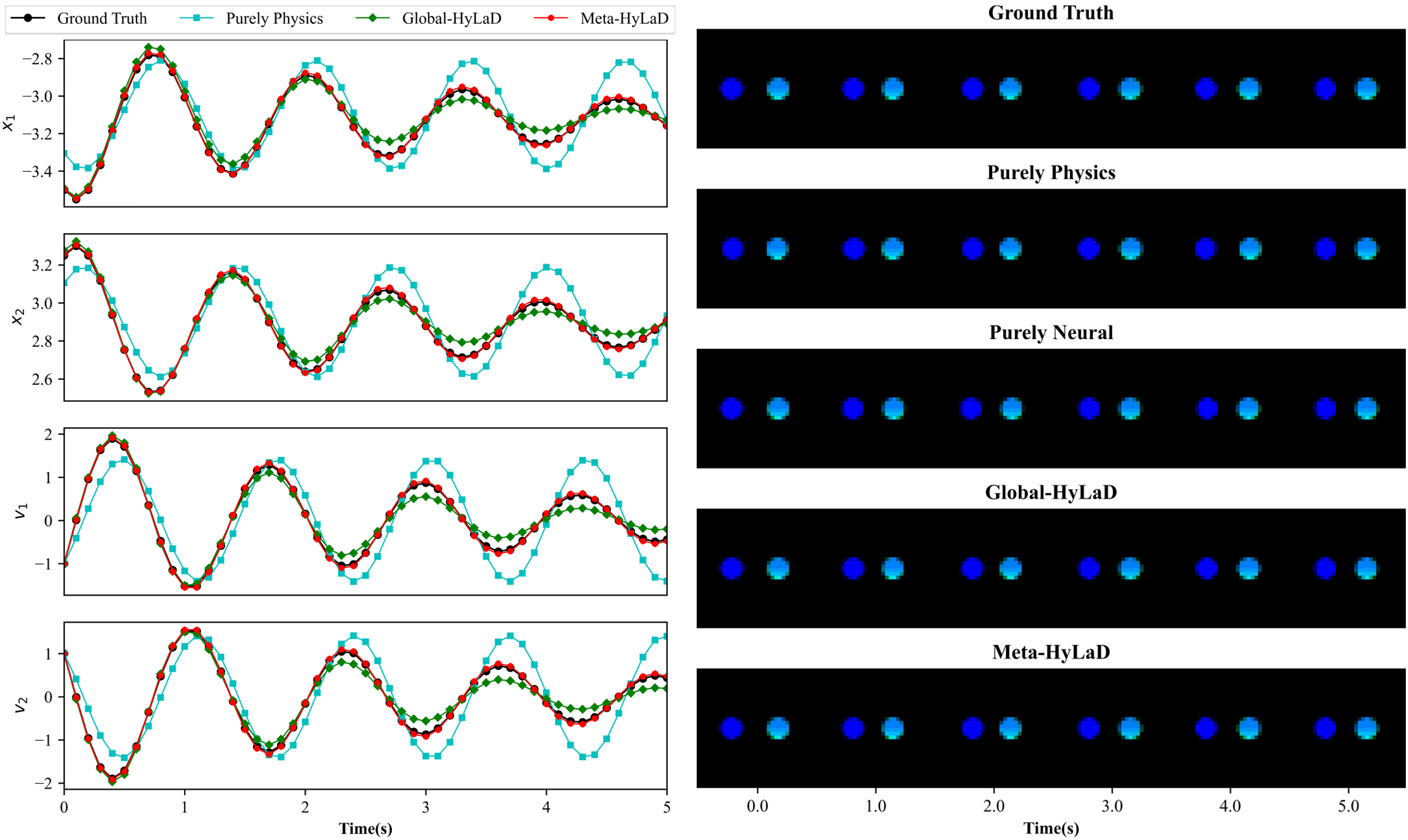}
    \caption{Visual examples of Mass Spring}
\end{figure}

\newpage
\subsection{Bouncing Ball}

\begin{table}[h]
\centering
\caption{Results of Boucing Ball}
\begin{tabular}{llcccc}
\hline
\multicolumn{1}{c}{Dynamics} & \multicolumn{1}{c}{Data} & MSE $\mathbf{x}_t$ (e$^{-2}$)$\downarrow$  & MSE $\mathbf{z}_t$ (e$^{-1}$) $\downarrow$ & VPT $\mathbf{x}_t$ $\uparrow$ & VPT of $\mathbf{z}_t$ $\uparrow$ \\ \hline
\multirow{2}{*}{\begin{tabular}[c]{@{}l@{}}Full Physics\\ (oracle)\end{tabular}}
& ID & 0.05(0.02) & 0.03(0.02) & 1.00(0.01) & 0.96(0.04) \\
& OOD & 0.14(0.06) & 0.14(0.09) & 0.93(0.04) & 0.84(0.06) \\ \hline
\multirow{2}{*}{Purely Physics} 
& ID & 3.20(0.10) & 15.54(0.16) & 0.09(0.02) & 0.04(0.02) \\ 
& OOD & 3.67(0.13) & 17.73(0.70) & 0.03(0.01) & 0.01(0.00) \\ \hline
\multirow{2}{*}{Purely Neural} 
& ID & 1.26(0.01) & 2.55(0.01) & 0.10(0.01) & 0.02(0.00) \\
& OOD & 1.92(0.23) & 7.26(2.98) & 0.12(0.02) & 0.03(0.01) \\ \hline
\multirow{2}{*}{Global-HyLaD} 
& ID & 2.15(0.04) & 8.33(0.19) & 0.21(0.03) & 0.12(0.02) \\ 
& OOD  & 3.24(0.04) & 15.21(0.32) & 0.06(0.00) & 0.03(0.00)  \\ \hline
\multirow{2}{*}{Meta-HyLaD} 
& ID & \textbf{0.06(0.02)} & \textbf{0.03(0.01)} & \textbf{1.00(0.00)} & \textbf{0.94(0.03)} \\ 
& OOD & \textbf{0.16(0.05)} & \textbf{0.17(0.07)} & \textbf{0.92(0.02)} & \textbf{0.80(0.05)} \\ \hline
\end{tabular}
\end{table}

\begin{figure}[h]
    \centering
    \includegraphics[width=0.95\linewidth]{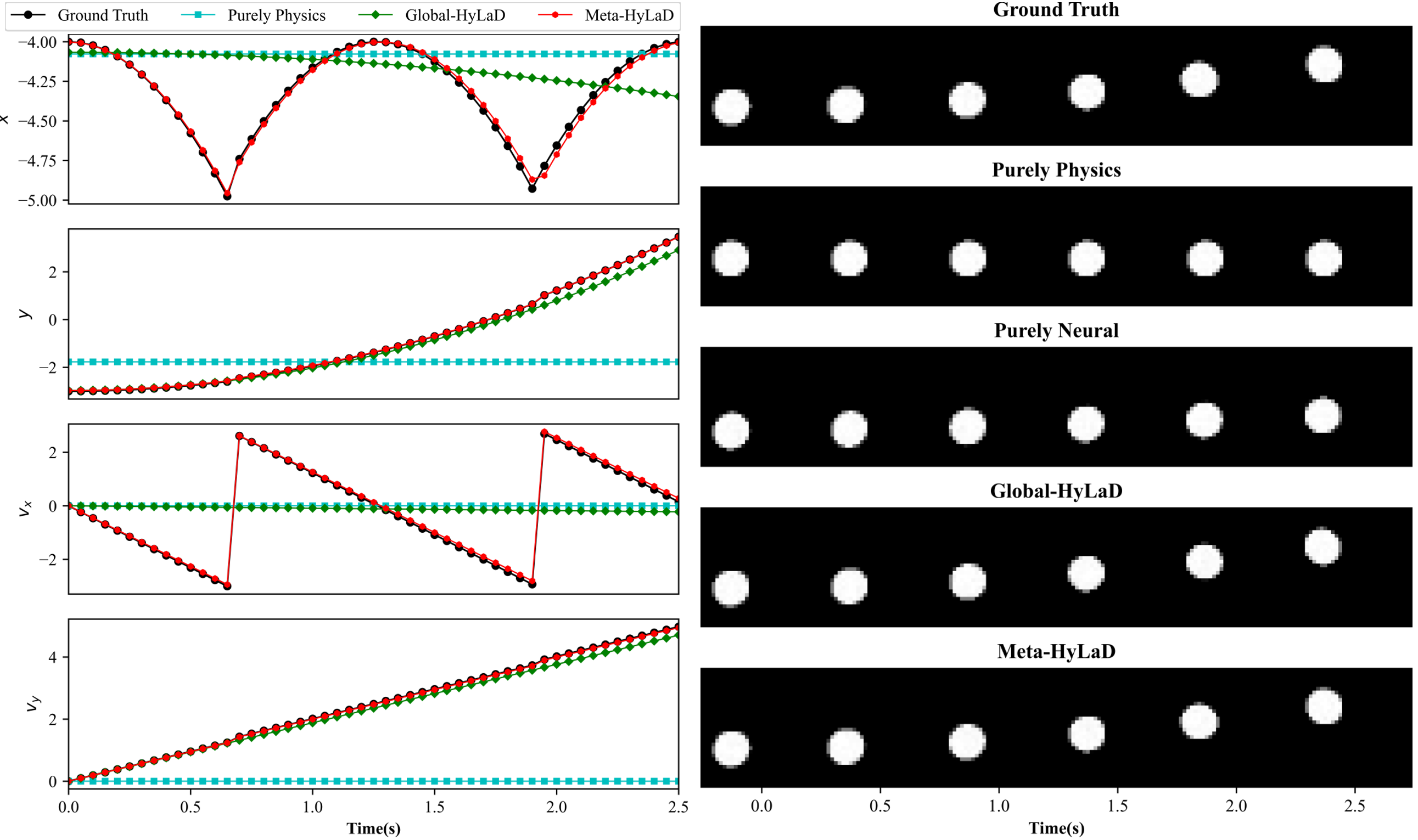}
    \caption{Visual examples of Bouncing Ball}
\end{figure}

\newpage
\subsection{Rigid Body}
\begin{table}[h]
\centering
\caption{Results of Rigid Body}
\begin{tabular}{llcccc}
\hline
\multicolumn{1}{c}{Dynamics} & \multicolumn{1}{c}{Data} & MSE $\mathbf{x}_t$ (e$^{-3}$)$\downarrow$  & MSE $\mathbf{z}_t$ (e$^{-2}$) $\downarrow$ & VPT $\mathbf{x}_t$ $\uparrow$ & VPT of $\mathbf{z}_t$ $\uparrow$ \\ \hline
\multirow{2}{*}{\begin{tabular}[c]{@{}l@{}}Full Physics\\ (oracle)\end{tabular}}
& ID & 0.37(0.19) & 0.02(0.01) & 1.00(0.01) & 1.00(0.00) \\
& OOD & 1.26(0.02) & 0.09(0.00) & 0.94(0.03) & 0.98(0.02) \\ \hline
\multirow{2}{*}{Purely Physics} 
& ID & 6.46(0.08) & 2.56(0.02) & 0.59(0.00) & 0.59(0.00) \\ 
& OOD & 18.23(0.23) & 10.36(0.24) & 0.29(0.00) & 0.38(0.00) \\ \hline
\multirow{2}{*}{Purely Neural} 
& ID & 5.01(0.17) & 15.92(0.93) & 0.58(0.01) & 0.00(0.00) \\
& OOD & 12.29(0.27) & 19.56(0.39) & 0.36(0.03) & 0.00(0.00) \\ \hline
\multirow{2}{*}{Global-HyLaD} 
& ID & 2.90(0.33) & 0.48(0.08) & 0.79(0.02) & 0.86(0.01) \\ 
& OOD & 14.41(0.92) & 5.28(1.00) & 0.34(0.02) & 0.46(0.02) \\ \hline
\multirow{2}{*}{Meta-HyLaD} 
& ID & \textbf{0.74(0.34)} & \textbf{0.05(0.04)} & \textbf{0.98(0.02)} & \textbf{0.99(0.01)} \\ 
& OOD & \textbf{1.74(0.12)} & \textbf{0.13(0.01)} & \textbf{0.90(0.01)} & \textbf{0.97(0.00)} \\ \hline
\end{tabular}
\end{table}

\begin{figure}[h]
    \centering
    \includegraphics[width=0.95\linewidth]{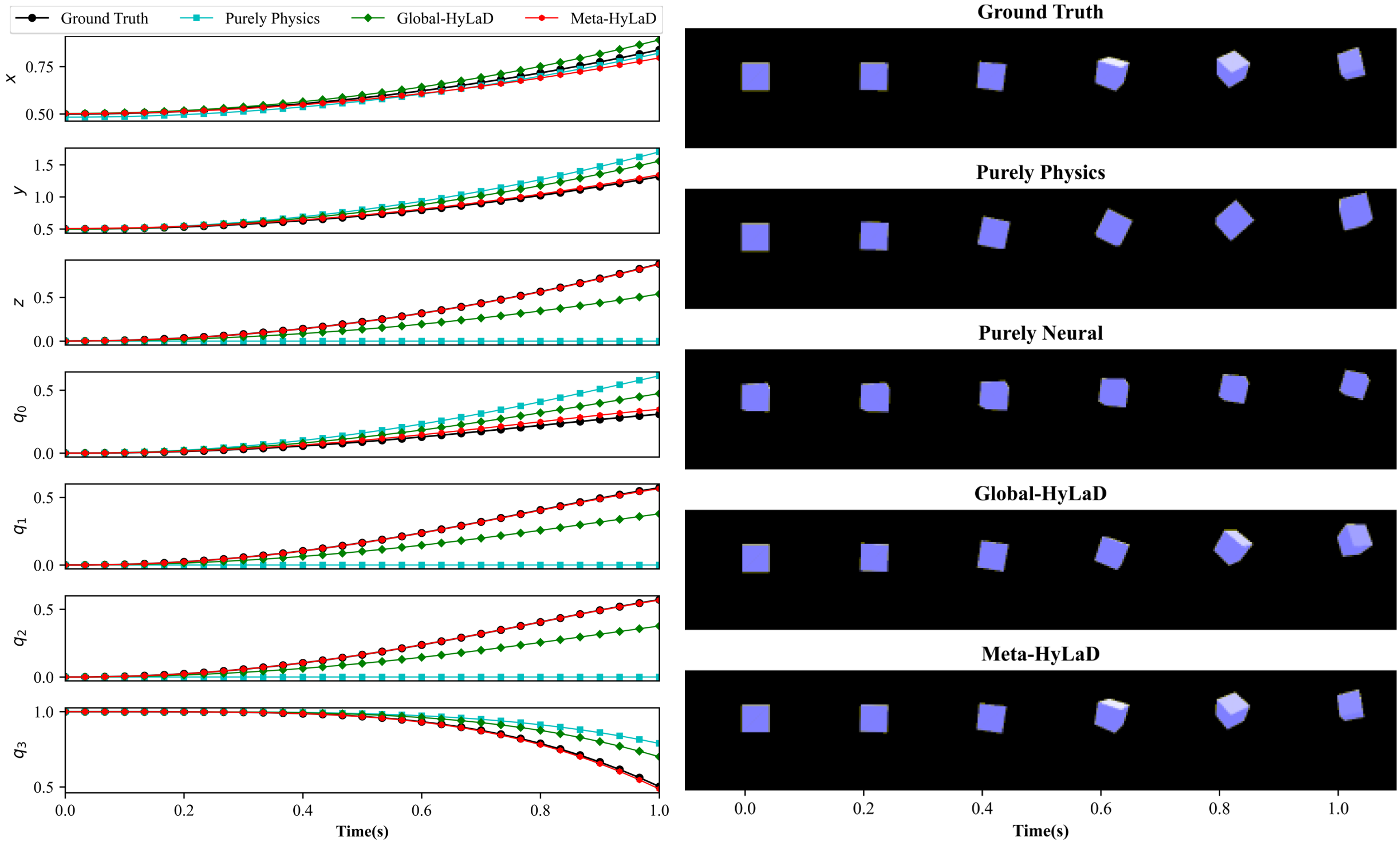}
    \caption{Visual examples of Rigid Body}
\end{figure}

\newpage
\subsection{Double Pendulum}
\begin{table}[h]
\centering
\caption{Results of Double Pendulum}
\begin{tabular}{llcccc}
\hline
\multicolumn{1}{c}{Dynamics} & \multicolumn{1}{c}{Data} & MSE $\mathbf{x}_t$ (e$^{-3}$)$\downarrow$  & MSE $\mathbf{z}_t$ (e$^{-1}$) $\downarrow$ & VPT $\mathbf{x}_t$ $\uparrow$ & VPT of $\mathbf{z}_t$ $\uparrow$ \\ \hline
\multirow{2}{*}{\begin{tabular}[c]{@{}l@{}}Full Physics\\ (oracle)\end{tabular}}
& ID & 0.18(0.01) & 0.13(0.01) & 1.00(0.00) & 0.99(0.00) \\
& OOD & 0.92(0.11) & 2.14(0.37) & 0.95(0.01) & 0.95(0.00) \\ \hline
\multirow{2}{*}{Purely Physics} 
& ID & 3.93(0.01) & 8.37(0.89) & 0.47(0.01) & 0.36(0.01) \\ 
& OOD  & 3.03(0.08) & 4.08(0.51) & 0.56(0.03) & 0.59(0.02) \\ \hline
\multirow{2}{*}{Purely Neural} 
& ID & 3.41(0.02) & 35.73(0.00) & 0.61(0.00) & 0.13(0.01) \\
& OOD & 4.10(0.17) & 37.35(0.00) & 0.57(0.04) & 0.14(0.01) \\ \hline
\multirow{2}{*}{Global-HyLaD} 
& ID & 1.64(0.01) & 2.79(0.03) & 0.86(0.00) & 0.78(0.00) \\ 
& OOD  & 2.84(0.17) & 6.55(0.26) & 0.72(0.03) & 0.75(0.01)  \\ \hline
\multirow{2}{*}{Meta-HyLaD} 
& ID & \textbf{0.52(0.05)} & \textbf{0.38(0.03)} & \textbf{0.98(0.00)} & \textbf{0.93(0.00)} \\ 
& OOD & \textbf{1.10(0.06)} & \textbf{2.97(0.19)} & \textbf{0.94(0.01)} & \textbf{0.95(0.01)} \\ \hline
\end{tabular}
\end{table}

\begin{figure}[h]
    \centering
    \includegraphics[width=0.95\linewidth]{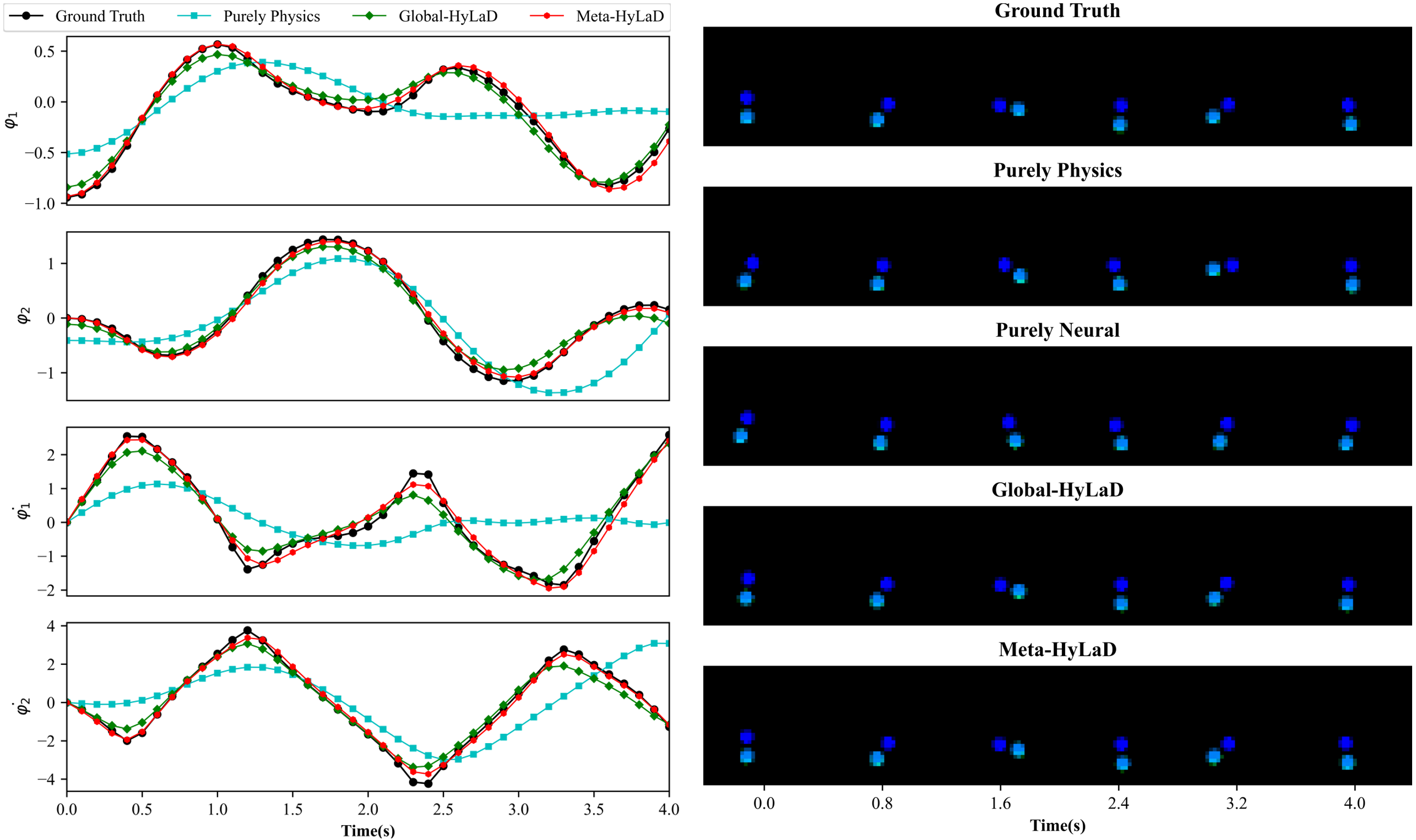}
    \caption{Visual examples of Double Pendulum}
\end{figure}

\newpage
\section{Complete Results of \cref{subsec:exp:baselines}: Comparison with Baselines}
Here we provide complete experimental results of \cref{subsec:exp:baselines}.
\label{appendix: result-sec4.4}
\begin{table}
\centering
\caption{Results of different baselines and Meta-HyLaD 1}
\label{Results} 
\resizebox{0.9\linewidth}{!}{
\begin{tabular}{|cl|cc|cc|cc|}
\hline
\multicolumn{8}{|c|}{Pendulum} \\ \hline
\multicolumn{1}{|l}{} & & \multicolumn{2}{c|}{Physics-based Decoder} & \multicolumn{2}{c|}{Neural Decoder with prior} & \multicolumn{2}{c|}{Neural Decoder no prior} \\ \hline
Model & Data & 
\multicolumn{1}{c}{MSE(e$^{-3}$)} & \multicolumn{1}{c|}{VPT} & \multicolumn{1}{c}{MSE(e$^{-3}$)} & \multicolumn{1}{c|}{VPT} & \multicolumn{1}{c}{MSE(e$^{-3}$)} & \multicolumn{1}{c|}{VPT} \\
\hline
\multirow{2}{*}{LSTM}    
& ID & 7.38(1.90) & 0.09(0.05) & / & / & 4.75(0.46) & 0.18(0.12) \\
& OOD & 5.36(0.74) & 0.10(0.03) & / & / & 4.09(0.19) & 0.19(0.12) \\ \hline
\multirow{2}{*}{Meta-LSTM}    
& ID & 7.13(1.13) & 0.04(0.02) & / & / & 5.13(0.01) & 0.04(0.00) \\
& OOD & 5.16(0.49) & 0.04(0.01)  & / & / & 3.68(0.01) & 0.04(0.00) \\ \hline
\multirow{2}{*}{NODE}    
& ID & 8.56(0.14) & 0.06(0.01)  &  / & / & 4.21(0.02) & 0.28(0.02) \\
& OOD & 5.16(0.44) & 0.07(0.01) & / & / & 3.85(0.11) & 0.30(0.03) \\ \hline
\multirow{2}{*}{Meta-NODE}    
& ID & 8.52(0.05) & 0.05(0.00)  & / & /  & 2.88(0.01) & 0.55(0.00)  \\
& OOD & 5.35(0.11) & 0.07(0.00)  & / & / & 3.50(0.10) & 0.43(0.01)  \\ \hline
\multirow{2}{*}{HGN}    
& ID & 6.47(0.17) & 0.11(0.03) & 3.53(0.02) & 0.48(0.01) & 2.99(0.03) & 0.62(0.01) \\
& OOD & 5.64(0.27) & 0.09(0.03) & 3.69(0.15) &0.37(0.02)  & 3.10(0.17) & 0.47(0.03) \\ \hline
\multirow{2}{*}{Meta-HGN}    
& ID & 9.69(0.15) & 0.03(0.00) & 3.04(0.17) & 0.54(0.01) & 0.67(0.08) & 0.98(0.00) \\
& OOD & 6.24(0.14) & 0.04(0.00) & 3.69(0.08) & 0.42(0.01) &0.72(0.05)  & 0.91(0.02) \\ \hline
\multirow{2}{*}{ALPS}    
& ID & 3.56(0.03) & 0.16(0.01) & 2.70(0.01) & 0.28(0.01) & 2.68(0.02) & 0.27(0.02) \\
& OOD & 4.17(0.00) & 0.04(0.00) & 3.28(0.01) & 0.06(0.00) & 3.25(0.03) & 0.06(0.00) \\ \hline
\multirow{2}{*}{Meta-HyLaD}    
& ID & \textbf{0.30(0.02)}  & \textbf{0.99(0.00)}  & \textbf{0.29(0.03)}  & \textbf{0.99(0.00)}  & \textbf{0.25(0.01)}  & \textbf{0.99(0.00)}  \\
& OOD & \textbf{0.31(0.06)}  & \textbf{0.99(0.00)}  & \textbf{0.25(0.02)}  & \textbf{0.99(0.00)}  & \textbf{0.20(0.04)}  & \textbf{1.00(0.00)}  \\ \hline
\multicolumn{8}{|c|}{Mass Spring} \\ \hline
\multicolumn{1}{|l}{} & & \multicolumn{2}{c|}{Physics-based Decoder} & \multicolumn{2}{c|}{Neural Decoder with prior} & \multicolumn{2}{c|}{Neural Decoder no prior} \\ \hline
Model & Data & 
\multicolumn{1}{c}{MSE(e$^{-3}$)} & \multicolumn{1}{c|}{VPT} & \multicolumn{1}{c}{MSE(e$^{-3}$)} & \multicolumn{1}{c|}{VPT} & \multicolumn{1}{c}{MSE(e$^{-3}$)} & \multicolumn{1}{c|}{VPT} \\
\hline
\multirow{2}{*}{LSTM}    
& ID & 16.59(12.95) & 0.01(0.01) & / & / & 0.40(0.03) & 0.56(0.06)   \\
& OOD & 16.51(13.11) & 0.01(0.01)   & / & / & 0.20(0.02) & 0.67(0.09) \\ \hline
\multirow{2}{*}{Meta-LSTM}    
& ID & 19.48(8.84) & 0.00(0.00)   & / & /  & 0.40(0.03) & 0.59(0.06)\\
& OOD & 19.44(8.94) & 0.00(0.00)    & / & / & 0.20(0.01) & 0.60(0.02) \\ \hline
\multirow{2}{*}{NODE}    
& ID & 0.80(0.01) & 0.15(0.01)    & / & / & 0.38(0.01) & 0.62(0.02) \\
& OOD & 0.38(0.01) & 0.25(0.01)    & / & / & 0.19(0.01) & 0.67(0.03) \\ \hline
\multirow{2}{*}{Meta-NODE}    
& ID & 0.69(0.09) & 0.21(0.05)    & / & / & 0.27(0.02) & 0.77(0.01)  \\
& OOD & 0.26(0.10) & 0.44(0.16)    & / & / & 0.24(0.01) & 0.69(0.02) \\ \hline
\multirow{2}{*}{HGN}    
& ID & 10.33(7.47) & 0.00(0.00) & 0.27(0.02) & 0.84(0.02) & 0.15(0.01) & 0.95(0.01) \\
& OOD & 10.11(7.60) & 0.23(0.03) & 0.11(0.01) & 0.78(0.04) & 0.11(0.01) & 0.85(0.03) \\ \hline
\multirow{2}{*}{Meta-HGN}    
& ID & 5.96(3.90) & 0.00(0.00) & 0.18(0.02) & 0.85(0.03) & 0.08(0.02) & 1.00(0.00) \\
& OOD & 5.66(3.96) & 0.00(0.00) & 0.18(0.03) & 0.76(0.02) & 0.06(0.01) & 0.99(0.00) \\ \hline
\multirow{2}{*}{ALPS}    
& ID & 17.52(7.60) & 0.00(0.00) & 2.59(0.80) & 0.03(0.03) & 1.40(0.59) & 0.10(0.10) \\
& OOD & 17.67(7.64) & 0.00(0.00) & 2.25(1.07) & 0.05(0.05) & 1.16(0.61) & 0.13(0.11) \\ \hline
\multirow{2}{*}{Meta-HyLaD}    
& ID & \textbf{0.01(0.00)}  & \textbf{1.00(0.00)}  & \textbf{0.03(0.00)}  & \textbf{1.00(0.00)} & \textbf{0.04(0.01)}  & 
\textbf{1.00(0.00)} \\
& OOD & \textbf{0.02(0.01)}  & \textbf{1.00(0.00)}  & \textbf{0.03(0.00)}  & \textbf{1.00(0.00)}  & \textbf{0.03(0.01)}  & \textbf{1.00(0.00)}  \\ \hline
\end{tabular}
}
\end{table}

\begin{table}
\centering
\caption{Results of different baselines and Meta-HyLaD 2}
\label{Results2}
\resizebox{0.9\linewidth}{!}{
\begin{tabular}{|cl|cc|cc|cc|}
\hline
\multicolumn{8}{|c|}{Bouncing Ball} \\ \hline
\multicolumn{1}{|l}{} & & \multicolumn{2}{c|}{Physics-based Decoder} & \multicolumn{2}{c|}{Neural Decoder with prior} & \multicolumn{2}{c|}{Neural Decoder no prior} \\ \hline
Model & Data & 
\multicolumn{1}{c}{MSE(e$^{-3}$)} & \multicolumn{1}{c|}{VPT} & \multicolumn{1}{c}{MSE(e$^{-3}$)} & \multicolumn{1}{c|}{VPT} & \multicolumn{1}{c}{MSE(e$^{-3}$)} & \multicolumn{1}{c|}{VPT} \\
\hline
\multirow{2}{*}{LSTM}    
& ID & 62.35(7.22) & 0.00(0.00) & / & / & 30.74(1.94) & 0.00(0.00)  \\
& OOD & 60.29(5.78) & 0.00(0.00) & / & / & 34.25(0.29) & 0.00(0.00)  \\ \hline
\multirow{2}{*}{Meta-LSTM}    
& ID & 79.03(9.22) & 0.01(0.00) & / & / & 26.37(0.93) & 0.00(0.00)  \\
& OOD & 76.66(11.14) & 0.01(0.00) & / & / & 35.84(2.29) & 0.00(0.00)   \\ \hline
\multirow{2}{*}{NODE}    
& ID & 38.74(0.17) & 0.03(0.00) & / & / & 22.33(1.13)  & 0.00(0.00)  \\
& OOD & 42.17(0.73) & 0.01(0.00) & / & / & 33.37(1.18) & 0.00(0.00)  \\ \hline
\multirow{2}{*}{Meta-NODE}    
& ID & 12.64(0.14) & 0.10(0.01) & / & / & 10.16(0.28) & 0.00(0.00)  \\
& OOD & 19.21(2.33) & 0.12(0.02) & / & / & 18.04(7.94) &  0.10(0.00) \\ \hline
\multirow{2}{*}{HGN}    
& ID & 33.92(3.69) & 0.01(0.00) & 15.65(7.04) & 0.06(0.02) & 11.41(3.04) & 0.15(0.05) \\
& OOD & 39.46(5.69) & 0.01(0.01) & 18.88(4.98) & 0.07(0.01) & 21.40(1.72) & 0.07(0.02) \\ \hline
\multirow{2}{*}{Meta-HGN}    
& ID & 18.70(6.67) & 0.02(0.01) & 3.87(0.27) & 0.70(0.04) & \textbf{3.48(0.97)} & \textbf{0.76(0.11)}\\
& OOD & 25.85(4.39) & 0.01(0.01) & 14.69(7.79) & 0.50(0.13) & 32.09(6.54) & 0.31(0.04) \\ \hline
\multirow{2}{*}{ALPS}    
& ID & 34.15(2.46) & 0.04(0.01) & 21.40(0.20) &  0.03(0.00) & 15.15(0.63) & 0.03(0.01) \\
& OOD & 33.95(1.22) & 0.01(0.00) & 24.92(1.12) & 0.02(0.01) & 23.54(4.06) & 0.02(0.01) \\ \hline
\multirow{2}{*}{Meta-HyLaD}    
& ID & \textbf{0.56(0.19)}  & \textbf{1.00(0.00)}  &\textbf{2.13(1.37)}  & \textbf{0.94(0.08)}  & 4.05(0.71) & 0.68(0.06) \\
& OOD & \textbf{1.60(0.50)}  & \textbf{0.92(0.02)}  & \textbf{4.46(1.25)}  & \textbf{0.81(0.04)}  & \textbf{5.92(1.22)}  & \textbf{0.61(0.02)}  \\ \hline
\multicolumn{8}{|c|}{Rigid Body} \\ \hline
\multicolumn{1}{|l}{} & & \multicolumn{2}{c|}{Physics-based Decoder} & \multicolumn{2}{c|}{Neural Decoder with prior} & \multicolumn{2}{c|}{Neural Decoder no prior} \\ \hline
Model & Data & 
\multicolumn{1}{c}{MSE(e$^{-3}$)} & \multicolumn{1}{c|}{VPT} & \multicolumn{1}{c}{MSE(e$^{-3}$)} & \multicolumn{1}{c|}{VPT} & \multicolumn{1}{c}{MSE(e$^{-3}$)} & \multicolumn{1}{c|}{VPT} \\
\hline
\multirow{2}{*}{LSTM}    
& ID & 21.43(3.99) & 0.02(0.03)   & / & / & 10.40(0.11) & 0.35(0.01) \\
& OOD & 19.63(5.15) & 0.02(0.04)  & / & / & 7.60(0.46)  & 0.43(0.02)\\ \hline
\multirow{2}{*}{Meta-LSTM}    
& ID & 18.80(4.85) & 0.06(0.05) & / & / & 3.97(0.69) & 0.72(0.11)   \\
& OOD & 16.66(6.27) & 0.08(0.07) & / & / & 5.58(1.77) & 0.60(0.15)   \\ \hline
\multirow{2}{*}{NODE}    
& ID & 17.73(6.42) & 0.22(0.18) & / & / & 9.29(0.01) & 0.42(0.00)   \\
& OOD & 15.55(8.24) & 0.27(0.23) & / & / & 6.33(0.07) & 0.50(0.00)   \\ \hline
\multirow{2}{*}{Meta-NODE}    
& ID & 4.65(1.56) & 0.60(0.11)   & / & /  & \textbf{1.51(0.08)} & \textbf{0.99(0.00)}  \\
& OOD & 5.00(1.98) & 0.58(0.15)  & / & / & \textbf{1.70(0.10)}  & \textbf{0.96(0.02)}  \\ \hline
\multirow{2}{*}{HGN}    
& ID & 11.76(3.42) & 0.32(0.01) & 2.65(0.13) & 0.85(0.02) & 6.36(0.66) & 0.61(0.02) \\
& OOD & 10.38(1.61) & 0.35(0.08) & 3.87(0.47) & 0.73(0.04) & 11.59(1.97) & 0.47(0.02) \\ \hline
\multirow{2}{*}{Meta-HGN}    
& ID & 6.77(1.56) & 0.61(0.02) &  \textbf{1.44(0.04)} & \textbf{1.00(0.00)}  & 5.45(0.22) & 0.62(0.03) \\
& OOD & 7.30(2.10) & \textbf{0.98(0.02)}  & \textbf{1.62(0.03)} & 0.98(0.01) & 5.53(1.03) & 0.62(0.08) \\ \hline
\multirow{2}{*}{ALPS}    
& ID & 5.03(0.57) & 0.54(0.05) & 2.53(0.18) & 0.92(0.03) & 5.77(1.13) & 0.54(0.12) \\
& OOD & 3.57(0.14) & 0.64(0.02) & 2.81(0.15) & 0.85(0.03) & 10.62(2.91) & 0.42(0.14) \\ \hline
\multirow{2}{*}{Meta-HyLaD}    
& ID &\textbf{0.74(0.34)}  & \textbf{0.98(0.02)}  & 1.75(0.10) & 0.99(0.00) & 5.11(0.22) & 0.63(0.02) \\
& OOD & \textbf{1.74(0.13)}  & 0.90(0.01) & 1.80(0.07) & \textbf{0.99(0.01)} & 7.69(2.08) & 0.47(0.08) \\ \hline
\multicolumn{8}{|c|}{Double Pendulum} \\ \hline
\multicolumn{1}{|l}{} & & \multicolumn{2}{c|}{Physics-based Decoder} & \multicolumn{2}{c|}{Neural Decoder with prior} & \multicolumn{2}{c|}{Neural Decoder no prior} \\ \hline
Model & Data & 
\multicolumn{1}{c}{MSE(e$^{-3}$)} & \multicolumn{1}{c|}{VPT} & \multicolumn{1}{c}{MSE(e$^{-3}$)} & \multicolumn{1}{c|}{VPT} & \multicolumn{1}{c}{MSE(e$^{-3}$)} & \multicolumn{1}{c|}{VPT} \\
\hline
\multirow{2}{*}{LSTM}    
& ID &  3.92(0.41) & 0.48(0.02) &/ & / &2.79(0.25) & 0.68(0.11) \\
& OOD & 4.64(0.28) & 0.43(0.01)    & / & / &  3.05(0.25) & 0.60(0.08) \\ \hline
\multirow{2}{*}{Meta-LSTM}    
& ID & 3.57(0.25) & 0.57(0.02)    & / & / & 2.32(0.05) & 0.83(0.02) \\
& OOD & 4.25(0.15) & 0.53(0.02)    & / & / & 2.59(0.06) & 0.76(0.02) \\ \hline
\multirow{2}{*}{NODE}    
& ID & 5.68(1.53) & 0.30(0.14)    & / & / & 1.89(0.13) & 0.90(0.01) \\
& OOD & 6.03(1.46) & 0.26(0.14)    & / & / & 2.42(0.08) & 0.78(0.01) \\ \hline
\multirow{2}{*}{Meta-NODE}    
& ID & 4.40(0.12) & 0.40(0.01)    & / & / & 1.37(0.06) & 0.97(0.00) \\
& OOD & 4.87(0.35) & 0.36(0.01)    & / & / & 2.05(0.05) & 0.84(0.01) \\ \hline
\multirow{2}{*}{HGN}    
& ID & 3.88(0.24) & 0.48(0.02) & 1.61(0.03) & 0.91(0.01) & 1.16(0.04) & 0.96(0.01) \\
& OOD & 4.57(0.33) & 0.44(0.01) & 2.25(0.10) & 0.78(0.01) & 1.91(0.16) & 0.84(0.03) \\ \hline
\multirow{2}{*}{Meta-HGN}    
& ID & 5.29(1.82) & 0.33(0.16) & \textbf{1.50(0.13)}  & \textbf{0.94(0.03)} & \textbf{0.64(0.03)}  & \textbf{1.00(0.00)}  \\
& OOD & 5.89(1.55) & 0.30(0.15) & 2.09(0.08) & 0.82(0.01) & 
\textbf{1.18(0.02)}  & \textbf{0.95(0.01)}  \\ \hline
\multirow{2}{*}{ALPS}    
& ID & 2.98(0.04) & 0.42(0.00) & 1.87(0.05) & 0.84(0.02) & 1.17(0.27) & 0.98(0.02) \\
& OOD & 2.47(0.10) & 0.67(0.01) & \textbf{1.99(0.06)} & \textbf{0.83(0.04)}  & 1.66(0.27) & 0.91(0.03) \\ \hline
\multirow{2}{*}{Meta-HyLaD}    
& ID & \textbf{0.52(0.05)}  & \textbf{0.98(0.00)}  & 1.63(0.11) & 0.92(0.03) & 1.15(0.11) & 0.98(0.01) \\
& OOD & \textbf{1.10(0.06)} & \textbf{0.94(0.01)}  & 2.24(0.06) & 0.80(0.00) & 1.77(0.14) & 0.89(0.02) \\ \hline
\end{tabular}
}
\end{table}

\newpage
\section{Data Details and Experimental Settings of Biomedical System: Dynamic PET`}
\label{appendix:data-PET}
In our experiment, we also consider a biomedical system: dynamic PET, which experiment can show that hybrid latent dynamics can learn a complex model with the help of a simple model. Here we provide more details of it.

\subsection{Background}Dynamic positron emission tomography (PET) imaging can provide measurements of spatial and temporal distribution of radiotracer-labeled biological substrates in living tissue\cite{PET}. Since Dynamic PET reconstruction is an ill-conditioned problem, the significance of incorporating prior knowledge into statistical reconstruction is well appreciated. The temporal kinetics of underlying physiological processes is an important physical prior in dynamic PET and various models have been proposed to model the tracer kinetics , which convert the radiotracer concentrations reconstructed from PET data into measures of the physiological processes. \\
\textbf{Compartment model: } Due to the simple implementation and biological plausibility, compartment models have been widely employed to quantitatively describe regional tracer kinetics in PET imaging where one need to postulate a linear or nonlinear structure in a number of compartments and their interconnections, and resolve them from the measurement data. Depending on the number of compartments and the complexity of the model, compartment model can be categorized into one-tissue compartment model, two-tissue compartment model and so on. In our experiment, we only consider one-tissue compartment model and two-tissue compartment model, which are written as:
\begin{align*}
    \text{one-tissue compartment model: }& \dot{C}_{Ti}(t)=-k_2C_{Ti}(t)+k_1C_P(t) \\
    \text{two-tissue compartment model: }& \begin{bmatrix}\dot{C}_{Ei}(t) \\ \dot{C}_{Mi}(t) \end{bmatrix}=\begin{bmatrix}-k_2-k_3 & k_4\\ k_3 & -k_4 \end{bmatrix}
                  \begin{bmatrix}C_{Ei}(t) \\ C_{Mi}(t)\end{bmatrix}+
                  \begin{bmatrix}k_1 \\ 0 \end{bmatrix}C_P(t)
\end{align*}
where $C_P(t)$: a space-invariant tracer delivery, $C_{Ei}(t), C_{Mi}(t)$: the concentration of radioactivity for any voxel $i(i=1,\dots,N)$ in different tissues(compartments) and $C_{Ti}(t)$: the total concentration of radioactivity in all tissues. For computational simplicity and without losing generality, we consider that 
$C_{Ti}(t)=C_{Ei}(t)+C_{Mi}(t)$ in two-tissue compartment model. And we assume the input $C_P(t)$ is known here, since $C_P(t)$ can be measured directly in practice. \\
\textbf{Imaging model: } Dynamic PET imaging involves a sequence of contiguous acquisition and a time series of activity images need to be reconstructed from the measurement data. For voxel $i(i=1,\dots,N)$, the $k$th scan $k=1,\dots,K$ attempts to measure the mean of the total concentration of radioactivity on the scanning time interval $[t_{k-1}, t_k]$, so the measured activity in scan $k$ for voxel $i$ is expressed as:
\begin{equation}
    x_{ik}(t)=\frac{1}{t_k-t_{k-1}}\int_{t_{k-1}}^{t_k}C_{Ti}(t)e^{-\frac{t}{\tau}}dt
\end{equation}
where $e^{-\frac{t}{\tau}}$ is the attenuation factor of radiotracer, The activity image of the $k$th scan $x_k$ is obtained by lexicographic ordering of the integrated radioactivity at different voxels $x_{ik}$. The raw data obtained by PET imaging device is sinogram $y_k=Dx_k+g_k$, where $D$ is the imaging matrix and $g_k$ is measurement errors. In our experiment, we simplify the original PET reconstruction problem by considering $x_k$ as the system observation and $C_T(t)$ as the system state $\mathbf{z}_t$.
\\

\begin{figure}[h]
    \centering
    \includegraphics[width=0.65\linewidth]{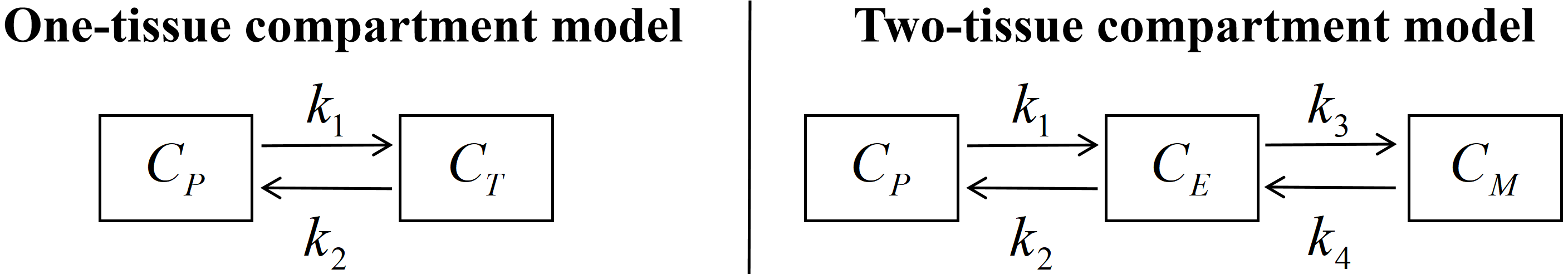}
    \caption{One-tissue and two-tissue compartment model in Dynamic PET}
    \label{Dynamic PET }
\end{figure}
\subsection{Data Details}
\label{Details:PET}
In experiment, we use the model proposed in \cite{Cp} for the input $C_P(t)$. 
\begin{equation}
    C_P(t)=(A_1t-A_2-A_3)e^{\lambda_1t}+A_2e^{\lambda_2t}+A_3e^{\lambda_3t}
\end{equation}
where $\lambda_1, \lambda_2$, and $\lambda_3 \text{ (in min}^{-1})$ are the eigenvalues of the model and $A_1$ (in $\mu$Ci/ml/min), $A2$ and $A3$ (in $\mu$Ci/ml) are the coefficients of the model. In experiment, we fix $A_2=21.8798, A_3=20.8113, \lambda_1=4.1339, \lambda_2=0.1191, \lambda_3=0.0104$ and $\tau=110$, then sample $A_1$ in range [100, 200]. And for two-tissue compartment model, we sample $k_1$ in range [0.1, 0.3], $k_2$ in range [0.01, 0.3], $k_3$ in range [0.01, 0.1] and $k_4$ in range [0.01, 0.05] according to \cite{DPET}. We generate concentration images $C_T(t)$ of a brain with 5 regions of interesting (ROIs) in time domain 0-40 min and we assume the pixels (voxels) in the same ROI share the same $k_1,k_2,k_3,k_4$. For activity images $o(t)$, the scanning period is 0.5 min, thus $k=80$. In total, we generate 12705 training and test sequences.
\subsection{Experimental Settings}
\begin{table}[h]
\centering
\resizebox{0.70\linewidth}{!}{
\begin{tabular}{ll}
\hline
\multicolumn{1}{c}{Dynamics} & \multicolumn{1}{c}{Equation} \\ \hline
Full Physics &  $\begin{bmatrix}\dot{C}_{Ei}(t) \\ \dot{C}_{Mi}(t) \end{bmatrix}=
\begin{bmatrix} \color{red}{-k_2-k_3} & \color{red}{k_4}\\ \color{red}{k_3} & \color{red}{-k_4} \end{bmatrix}
\begin{bmatrix}C_{Ei}(t) \\ C_{Mi}(t)\end{bmatrix}+
\begin{bmatrix}{\color{blue}{k_1}} \\ 0 \end{bmatrix}C_P(t)$
\\
Purely Physics & $\dot{C}_{Ti}(t)=-{\color{blue}{k_2}}C_{Ti}(t)+{\color{blue}{k_1}}C_P(t)$
\\
Purely Neural & $\dot{C}_{Ti}(t)=f_{\text{NN}_\phi}(C_{Ti}(t);\mathbf{c}_n)$
\\
Global-HyLaD  & $\dot{C}_{Ti}(t)=-{\color{blue}{k_2}}C_{Ti}(t)+{\color{blue}{k_1}}C_P(t)+f_{\text{NN}_\phi}(C_{Ti}(t))$
\\ 
Meta-HyLaD & $\dot{C}_{Ti}(t)=-{\color{blue}{k_2}}C_{Ti}(t)+{\color{blue}{k_1}}C_P(t)+f_{\text{NN}_\phi}(C_{Ti}(t);\mathbf{c}_n)$
\\ \hline
\end{tabular}
}
\end{table}

\subsection{Experimental Results}
\begin{table}[h]
\centering
\caption{Results of Dynamic PET}
\resizebox{0.70\linewidth}{!}{
\begin{tabular}{lcccc}
\hline
\multicolumn{1}{c}{Dynamics} & MSE of $o_t$ $\downarrow$  & MSE of $x_t$ $\downarrow$ & PSNR of $o_t$ $\uparrow$ & VPT-MSE of $x_t$ $\uparrow$ \\ \hline
Full Physics & 9.34(1.66)e-3 & 1.07(0.26)e-1 & 58.00(0.83) & 0.99(0.01) \\ 
Purely Physics & 1.08(0.01)e0 & 5.36(0.06)e0 & 36.84(0.25) & 0.00(0.00) \\ 
Purely Neural  & 2.77(0.35)e-1 & 1.50(0.32)e0 & 43.32(0.58) & 0.30(0.08) \\ 
Globa-HyLaD & 8.83(0.75)e-2 & 4.66(0.50)e-1 & 48.26(0.41) & 0.90(0.05)\\ 
Meta-HyLaD & \textbf{1.37(0.15)e-2} & \textbf{1.13(0.18)e-1} & \textbf{56.39(0.50)} & \textbf{1.00(0.00)} \\ \hline
\end{tabular}}
\end{table}

\begin{figure}[h]
    \centering
    \includegraphics[width=0.75\linewidth]{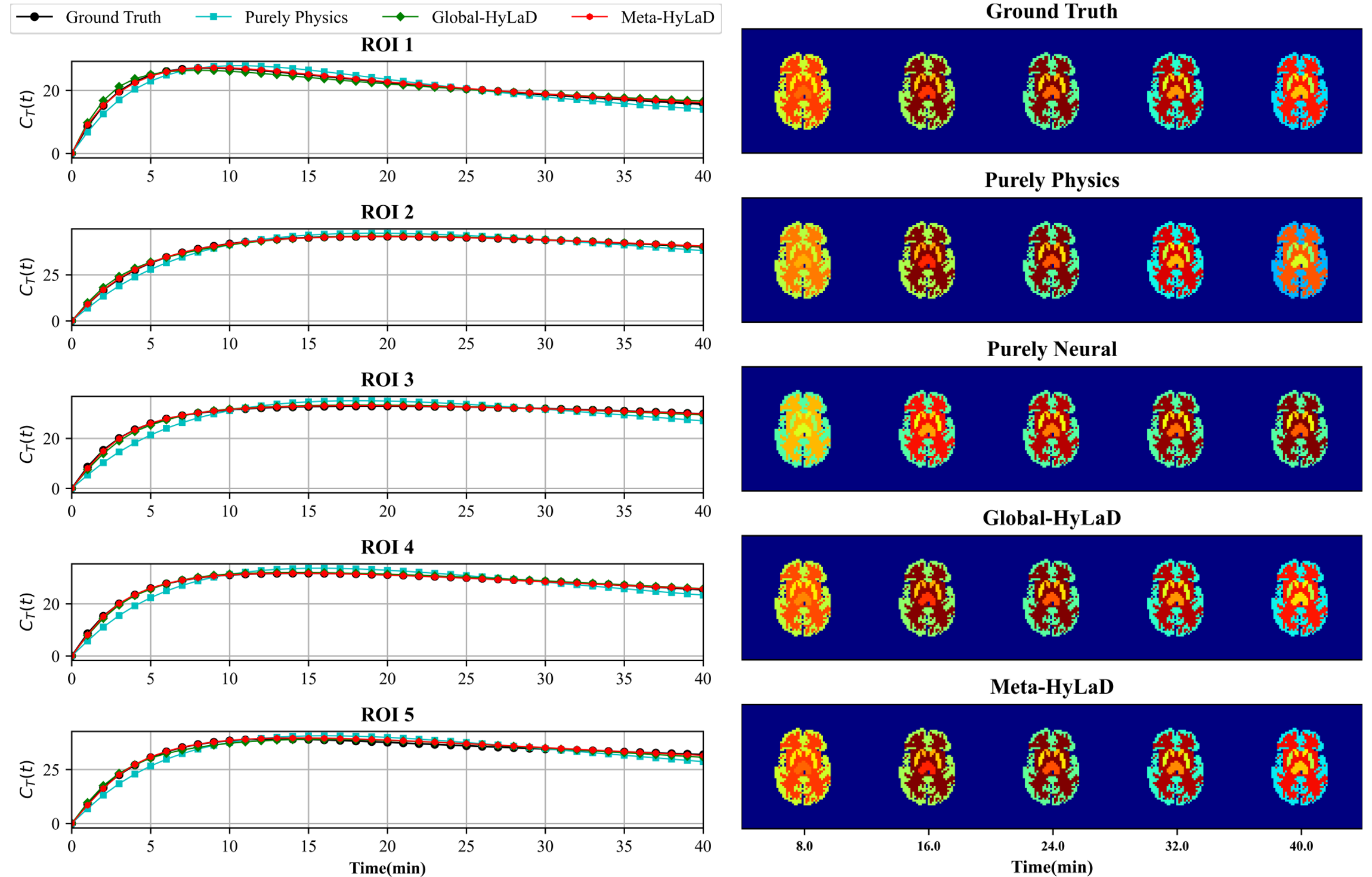}
    \caption{Dynamic PET}
\end{figure}

\newpage
\section{Ablation Study}
\label{appendix: result-sec4.6}
\subsection{The effect of the size of context set}
\label{app:subsec:k}
We test the effect of k on k-shot context set on two cases: 1) training and testing on the same and fixed k. 2) training on the variable k, but testing on fixed or variable k.

\begin{table}[h]
\centering
\caption{Training and Testing on the same and fixed k}
\resizebox{0.70\linewidth}{!}{
\begin{tabular}{clcccc}
\hline
\multicolumn{1}{c}{K} & MSE of $\mathbf{x}_t$(e-4) $\downarrow$  & MSE of $\mathbf{z}_t$(e-3) $\downarrow$ & VPT-MSE of $\mathbf{x}_t$ $\uparrow$ & VPT-MSE of $\mathbf{z}_t$ $\uparrow$ \\ \hline
1 & 4.90(0.41) & 6.08(0.12) & 0.98(0.00) & 0.96(0.01) \\
\hline
3 & 3.76(0.21) & 3.85(0.92) & 0.98(0.00) & 0.97(0.00) \\ 
\hline
5 & 3.35(0.21) & 2.53(0.09) & 0.99(0.00) & 0.98(0.00) \\
\hline
7 & \textbf{2.99(0.20)} & 2.24(0.20) & 0.99(0.00) & 0.98(0.00) \\ 
\hline
9 & 3.09(0.41) & \textbf{2.21(0.38)} & \textbf{0.99(0.00)} & \textbf{0.99(0.00)} \\ 
\hline
\end{tabular}}
\end{table}

\begin{table}[h]
\centering
\caption{Training on variable k between [1, 9], but testing on fixed or variable k}
\resizebox{0.70\linewidth}{!}{
\begin{tabular}{clcccc}
\hline
K & \multicolumn{1}{c}{Mode} & MSE of $\mathbf{x}_t$(e-4) $\downarrow$ & MSE of $\mathbf{z}_t$(e-3) $\downarrow$ &  VPT-MSE of $\mathbf{x}_t$ $\uparrow$ & VPT-MSE of $\mathbf{z}_t$ $\uparrow$ \\ \hline
1 & Fixed & 7.14(1.17) & 8.72(2.24) & 0.95(0.01) & 0.91(0.02)    \\ \hline
3 & Fixed & 4.79(0.39) & 4.97(0.72) & 0.98(0.01) & 0.96(0.01)    \\ \hline
5 & Fixed & 3.98(0.38) & 3.37(0.35) & 0.98(0.00) & 0.97(0.01)     \\ \hline
7 & Fixed & 3.43(0.35) & 2.64(0.22) & 0.99(0.00) & 0.98(0.00)    \\ \hline
\multirow{2}{*}{9} 
& Fixed     & \textbf{3.12(0.28)} & \textbf{2.27(0.19)} & \textbf{0.99(0.00)}  & \textbf{0.98(0.00)} \\
& Variable  & 3.88(0.23) & 3.50(0.14) & 0.99(0.00) & 0.97(0.00) \\ \hline
\end{tabular}}
\end{table}

\subsection{The robustness to different level of noise}
\label{app:subsec:noise}
We test the rubustness of Meta-HyLaD to different level of noise on observed time-series. The SNR of observed time-series is in range [5dB, 30dB].
\begin{table}[h]
\caption{The robustness to different level of noise}
\centering
\resizebox{0.70\linewidth}{!}{
\begin{tabular}{ccccc}
\hline
\multicolumn{1}{c}{SNR} & MSE of $\mathbf{x}_t$(e-4) $\downarrow$  & MSE of $\mathbf{z}_t$(e-3) $\downarrow$ & VPT-MSE of $\mathbf{x}_t$ $\uparrow$ & VPT-MSE of $\mathbf{z}_t$ $\uparrow$ \\ \hline
5 dB   & 4.02(0.17) & 3.33(0.16) & 0.98(0.00) & 0.96(0.00) \\
\hline
10 dB  & 3.89(0.47) & 3.09(0.47) & 0.99(0.00) & 0.97(0.01) \\
\hline
15 dB  & 3.44(0.06) & 2.66(0.07) & 0.99(0.00) & 0.98(0.00) \\ 
\hline
20 dB  & 3.51(0.55) & 2.61(0.54) & 0.99(0.00) & 0.98(0.01) \\ 
\hline
30 dB  & \textbf{3.41(0.04)} & \textbf{2.59(0.01)} & \textbf{0.99(0.00)}  & \textbf{0.99(0.00)}  \\ 
\hline
\end{tabular}}
\end{table}

\subsection{Generality of Meta-HyLaD}
\label{app:subsec:generality}
We investigate different choices of modeling 
the neural component $f_{\text{NN}_\phi}$ in Meta-HyLaD. As an example, on Pendulum, 
 we compare two types of Meta-HyLaD: Meta-HyLaD with prior where the input to $f_{\text{NN}_\phi}$ is selected to be the state variable $\dot{\varphi}$ as informed by physics,  
 and a more general version of Meta-HyLaD (Meta-HyLaD no prior) where the complete state variables are input to $f_{\text{NN}_\phi}$. 
\begin{equation}
    \begin{aligned}
        &\text{Meta-HyLaD with prior: }\frac{\mathrm{d}}{\mathrm{d} t}\begin{bmatrix} \varphi  \\ \dot{\varphi}  \end{bmatrix}= \begin{bmatrix} \dot{\varphi}  \\ -\frac{G}{L}\sin(\varphi) \end{bmatrix}+f_{\text{NN}_\phi}(\dot{\varphi}; \mathbf{c}_n) \\
        &\text{Meta-HyLaD no prior: }\frac{\mathrm{d}}{\mathrm{d} t}\begin{bmatrix} \varphi  \\ \dot{\varphi}  \end{bmatrix}= \begin{bmatrix} \dot{\varphi}  \\ -\frac{G}{L}\sin(\varphi) \end{bmatrix}+f_{\text{NN}_\phi}(\varphi,\dot{\varphi}; \mathbf{c}_n)
    \end{aligned}
\end{equation}
Similar examples are tested on Mass Spring:
\begin{equation}
    \begin{aligned}
        &\text{Meta-HyLaD with prior: } \frac{\mathrm{d}}{\mathrm{d} t}\begin{bmatrix} v_1  \\ v_2  \end{bmatrix}=
\begin{bmatrix} -\frac{\Tilde{x}}{\left | \Tilde{x} \right |} \frac{k}{m_1}
\left ( \left | \Tilde{x} \right | -l_0 \right ) \\ \frac{\Tilde{x}}{\left | \Tilde{x} \right |} \frac{k}{m_2}
\left ( \left | \Tilde{x} \right | -l_0 \right )
\end{bmatrix}+f_{\text{NN}_\phi}(v_1,v_2;\mathbf{c}_n) \\
        &\text{Meta-HyLaD no prior: }\frac{\mathrm{d}}{\mathrm{d} t}\begin{bmatrix} v_1  \\ v_2  \end{bmatrix}=
\begin{bmatrix} -\frac{\Tilde{x}}{\left | \Tilde{x} \right |} \frac{k}{m_1}
\left ( \left | \Tilde{x} \right | -l_0 \right ) \\ \frac{\Tilde{x}}{\left | \Tilde{x} \right |} \frac{k}{m_2}
\left ( \left | \Tilde{x} \right | -l_0 \right )
\end{bmatrix}+f_{\text{NN}_\phi}(x_1, x_2, v_1,v_2;\mathbf{c}_n)
    \end{aligned}
\end{equation}

 Results below show that the forecasting and identification performance of Meta-HyLaD are minimally affected by a more general expression of $f_{\text{NN}_\phi}$. 

\begin{table}[t]
\centering
\caption{Pendulum}
\begin{tabular}{lccc}
\hline
 & MSE $\mathbf{x}_t(\text{e}^{-4})\downarrow$ & MSE $\varphi_t(\text{e}^{-3})\downarrow$ & MSE $ \dot{\varphi_t}(\text{e}^{-3})\downarrow$ \\ \hline
Meta-HyLaD with prior & 2.96(0.15) & 1.05(0.12) & 3.40(0.19) \\ \hline
Meta-HyLaD no prior & 3.19(0.11) & 1.07(0.10) & 3.74(0.03) \\ \hline
\end{tabular}
\end{table}

\begin{table}[t]
\centering
\caption{Mass Spring}
\begin{tabular}{lccc}
\hline
 & MSE $\mathbf{x}_t(\text{e}^{-5})\downarrow$ & MSE $\varphi_t(\text{e}^{-4})\downarrow$ & MSE $ \dot{\varphi_t}(\text{e}^{-3})\downarrow$ \\ \hline
Meta-HyLaD with prior & 1.29(0.31) & 1.21(0.37) & 1.66(0.41) \\ \hline
Meta-HyLaD no prior & 1.19(0.16) & 1.20(0.00) & 1.74(0.13)\\ \hline
\end{tabular}
\end{table}

\subsection{Failure Modes for Meta-HyLaD}
\label{app:subsec:failure_hylad}

Here we probe the potential failure models for Meta-HyLaD, in particularly concerning the strength of the prior physics. 
On Pendulum, we compare 
two Meta-HyLaD formulations:
one with stronger prior physics as used in the main experiments, 
and one with weaker prior physics that is only aware of the dampening effect in the true physics. 
The results show that,
when the prior physics is too weak, 
Meta-HyLaD will approach the performance of a fully neural model at the data space, but still with significantly better results for the latent state variables.

\begin{equation}
    \begin{aligned}
        &\text{Meta-HyLaD with strong physics: }\frac{\mathrm{d}}{\mathrm{d} t}\begin{bmatrix} \varphi  \\ \dot{\varphi}  \end{bmatrix}= \begin{bmatrix} \dot{\varphi}  \\ -\frac{G}{L}\sin(\varphi) \end{bmatrix}+f_{\text{NN}_\phi}(\dot{\varphi}; \mathbf{c}_n) \\
        &\text{Meta-HyLaD with weak physics: }\frac{\mathrm{d}}{\mathrm{d} t}\begin{bmatrix} \varphi  \\ \dot{\varphi}  \end{bmatrix}= \begin{bmatrix} \dot{\varphi}  \\ -\beta \dot{\varphi} \end{bmatrix}+f_{\text{NN}_\phi}(\varphi; \mathbf{c}_n)
    \end{aligned}
\end{equation}

\begin{table}[t]
\centering
\caption{Results}
\begin{tabular}{lcccc}
\hline
 & MSE $\mathbf{x}_t(\text{e}^{-4})\downarrow$ & MSE $\mathbf{z}_t(\text{e}^{-2})\downarrow$ & VPT $\mathbf{x}_t\uparrow$ & VPT $\mathbf{z}_t\uparrow$ \\ \hline
Meta-HyLaD with strong physics & 2.96(0..06) & 2.16(0.03) & 0.99(0.00) & 0.98(0.00) \\ \hline
Meta-HyLaD with weak physics & 33.32(0.58) & 44.94(1.08) & 0.48(0.01) &  0.39(0.00) \\ \hline
Purely Neural &  33.55(5.18) & 167.06(106.23) & 0.53(0.07) & 0.00(0.00)   \\ \hline
\end{tabular}
\end{table}

\subsection{Failure Modes for Neural Decoder \& Mitigation by Meta-HyLaD}
\label{app:subsec:failure_decoder}
Given the strong performance of neural decoder as observed in \cref{subsec:exp:baselines}, here we probe its potential failure modes, 
especially considering the possibility that 
the true emission functions are not global but also change with each data sample. 

 As an example, we consider another experimental setting of Pendulum: $\frac{\mathrm{d}}{\mathrm{d} t}\begin{bmatrix} \varphi  \\ \dot{\varphi}  \end{bmatrix}= \begin{bmatrix} \dot{\varphi}  \\ -\frac{G}{{\color{blue}{L}}}\sin(\varphi)-{{\color{red}{\beta}}} \dot{\varphi} \end{bmatrix}$, we fix $G=10$ and ample $L\in[1.0, 3.0]$ with other details identical \cref{appendix:data-physics}. 
Because the true emission function 
relies on parameter $L$, 
this creates a scenario where a \textit{global} decoder 
can fail -- as shown in the results below from the standard neural decoder. 

Interestingly, 
if we simply provide the estimated physics parameter $c_p$ (here representing $L$) to the decoder (adaptive neural decoder), 
this challenge can be significantly reduced. 
While not perfectly addressing the issue, this leaves an interesting future avenue for Meta-HyLaD and its learn-to-identify learning strategies. 

\begin{equation}
    \begin{aligned}
    &\text{Standard neural decoder}:  \hat{\mathbf{x}}_t=g_{\psi}(\dot{\varphi}, L) \\
    & \text{Adaptive neural decoder}:  \hat{\mathbf{x}}_t=g_{\psi}(\dot{\varphi})
    \end{aligned}
\end{equation}
\begin{table}t]
\centering
\caption{Results}
\begin{tabular}{lcccc}
\hline
 & MSE $\mathbf{x}_t(\text{e}^{-4})\downarrow$ & MSE $\mathbf{z}_t(\text{e}^{-2})\downarrow$ & VPT $\mathbf{x}_t\uparrow$ & VPT $\mathbf{z}_t\uparrow$ \\ \hline
Physics-based decoder & 1.93(0.31) & 0.17(0.03) & 1.00(0.00) & 0.97(0.01) \\ \hline
Standard neural decoder & 39.24(2.95) & 50.57(2.26) & 0.27(0.01) &  0.01(0.00) \\ \hline
Adaptive neural decoder & 7.45(3.93) & 10.88(2.79) & 0.89(0.05) & 0.62(0.05) \\  \hline
\end{tabular}
\end{table}
\section{Implementation Details}
\label{app:implmentation}
In this section, we give more implementation details on each experiment over all modules and models. All experiments were run on NVIDIA TITAN RTX with 24 GB memory. We use Adam as optimizer with learning rate 1 × 10$^{-3}$.
\subsection{Architecture for Meta-HyLaD}
As shown in \cref{sec:method}, Meta-HyLaD has following basic modules: 1) Neural Dynamic Function $f_{\text{NN}_\phi}$: $\frac{d \mathbf{z}_t}{dt} = f_{\textrm{PHY}}(\mathbf{z}_{t};\mathbf{c}_p) + f_{\textrm{NN}_\phi}(\mathbf{z}_{t};\mathbf{c}_n)$; 2) Hyper Network: $\phi = h_\theta(\mathbf{c}_n)$; 3) Initial Encoder $\mathcal{E}_{\phi_z}$: $\hat{\mathbf{z}}_0 = \mathcal{E}_{\phi_z} (\mathbf{x}_{0:l})$; 4)
Neural Context Encoder $\mathcal{E}_{\zeta_n}$: $\hat{\mathbf{c}}_{n,j} = \frac{1}{k}{\textstyle \sum_{\mathbf{x}_{0:T}^s\in \mathcal{D}_j^s}} \mathcal{E}_{\zeta_n} (\mathbf{x}_{0:T}^{s})$; 5) Physical Context Encoder $\mathcal{E}_{\zeta_p}$: $\hat{\mathbf{c}}_{p,j}=\frac{1}{k}{\textstyle \sum_{\mathbf{x}_{0:T}^s\in \mathcal{D}_j^s}} \mathcal{E}_{\zeta_p} (\mathbf{x}_{0:T}^{s})$ or $\mathcal{E}_{\zeta_p} (\mathbf{x}_{0:T}^q)$; 6) Neural Decoder $g_{\zeta_D}$: $\hat{\mathbf{x}}_t=g_{\zeta_D}(\hat{\mathbf{z}}_t)$.
All these modules are composed of convolutional layers(CNN) and Fully Connected Layers(FNN) Here we provide more details about them.\\

\textbf{(1) Neural Dynamic Function $f_{\text{NN}_\phi}$: } 
\begin{lstlisting}
nn.Sequential(
    nn.Linear(input_dim, 8),
    nn.SiLU(),
    nn.Linear(8, 8),
    nn.SiLU(),
    nn.Linear(8, 8),
    nn.SiLU(),
    nn.Linear(8, 8),
    nn.SiLU(),
    nn.Linear(8, output_dim))
\end{lstlisting}

\textbf{(2) Hyper Network: $\phi = h_\theta(\mathbf{c}_n)$: } 
\begin{lstlisting}
nn.Sequential(
    nn.Linear(input_dim, 16),
    nn.SiLU(),
    nn.Linear(16, 16),
    nn.SiLU(),
    nn.Linear(16, 16),
    nn.SiLU(),
    nn.Linear(16, 8),
    nn.SiLU(),
    nn.Linear(8, output_dim))
\end{lstlisting}

\textbf{(3) Initial Encoder $\mathcal{E}_{\phi_z}$: } 
\begin{lstlisting}
nn.Sequential(
    nn.Conv2d(time_steps, num_filters, kernel_size=5, stride=2, padding=(2, 2)),
    nn.BatchNorm2d(num_filters),
    nn.LeakyReLU(0.1),
    nn.Conv2d(num_filters, num_filters * 2, kernel_size=5, stride=2, padding=(2, 2)),
    nn.BatchNorm2d(num_filters * 2),
    nn.LeakyReLU(0.1),
    nn.Conv2d(num_filters * 2, num_filters, kernel_size=5, stride=2, padding=(2, 2)),
    nn.Tanh(),
    Flatten())
nn.Linear(num_filters*16, output_dim)
\end{lstlisting}
\textbf{(4) Neural Encoder $\mathcal{E}_{\zeta_n}$: }
\begin{lstlisting}
class SpatialTemporalBlock(nn.Module):
    def __init__(self, t_in, t_out, n_in, n_out, last)
    self.conv = nn.Conv2d(n_in, n_out, kernel_size=5, stride=2, padding=(2,2))
    self.bn = nn.BatchNorm2d(n_out)
    self.act = nn.LeakyReLU(0.1)
    self.lin_t = nn.Linear(t_in, t_out)
nn.Sequential(
    SpatialTemporalBlock(time_steps, time_steps//2, 1, num_filters, False),
    SpatialTemporalBlock(time_steps//2, time_steps//4, num_filters, num_filters*2, False),
    SpatialTemporalBlock(time_steps//4, 1, num_filters*2, num_filters, True),
    Flatten())
nn.Linear(num_filters*16, output_dim)
\end{lstlisting}
\textbf{(5) Physical Context Encoder $\mathcal{E}_{\zeta_p}$: }
\begin{lstlisting}
nn.Sequential(
    SpatialTemporalBlock(time_steps, time_steps//2, 1, num_filters, False),
    SpatialTemporalBlock(time_steps//2, time_steps//4, num_filters, num_filters*2, False),
    SpatialTemporalBlock(time_steps//4, 1, num_filters*2, num_filters, True),
    Flatten())
nn.Linear(num_filters*16, latent_dim)
\end{lstlisting}
\textbf{(6) Neural Decoder $g_{\zeta_D}$: $\hat{\mathbf{x}}_t=g_{\zeta_D}(\hat{\mathbf{z}}_t)$: }
\begin{lstlisting}
nn.Sequential(
    nn.Linear(input_dim, 32),
    nn.ReLU(),
    nn.Linear(32, 32*4),
    nn.ReLU(),
    nn.Linear(32*4, 32*16),
    nn.ReLU(),
    nn.Linear(32*16, output_dim*output_dim))
\end{lstlisting}
\subsection{Architecture for Other Baselines}
\subsubsection{LSTM, Meta-LSTM, NODE, and Meta-NODE}
LSTM, Meta-LSTM, NODE, and Meta-NODE are modified from the code in \cite{jiang2022sequential}. Their modules are almost the same as Meta-HyLaD, only the neural dynamic function are different.\\
\textbf{(1) Neural Dynamic Function of LSTM and Meta-LSTM:}
\begin{lstlisting}
nn.Linear(input_dim, latent_dim)
nn.LSTMCell(input_size=latent_dim, hidden_size=transition_dim)
nn.Linear(transition_dim, output_dim)
\end{lstlisting}
\textbf{(2) Neural Dynamic Function of NODE and Meta-NODE:}
\begin{lstlisting}
nn.Sequential(
    nn.Linear(input_dim, 8),
    nn.SiLU(),
    nn.Linear(8, 8),
    nn.SiLU(),
    nn.Linear(8, 8),
    nn.SiLU(),
    nn.Linear(8, 8),
    nn.SiLU(),
    nn.Linear(8, output_dim))
\end{lstlisting}
\subsubsection{HGN and Meta-HGN}
HGN and Meta-HGN are modified from the code in \cite{HGN}. They have HamiltonNet in addition to Meta-HyLaD and neural dynamics are followed Hamiltonian mechanics. \\
\textbf{(1) HamiltonNet:}
\begin{lstlisting}
nn.Sequential(
    nn.Linear(input_dim, 64),
    nn.functional.softplus(),
    nn.Linear(64, 64),
    nn.functional.softplus(),
    nn.Linear(64, output_dim))
\end{lstlisting}
\textbf{(2) Neural dynamics followed Hamiltonian mechanics:}
\begin{lstlisting}
energy = HamiltonNet(q, p)
dq_dt = torch.autograd.grad(energy, p, 
        create_graph=True, retain_graph=True, grad_outputs=torch.ones_like(energy))[0]
dp_dt = -torch.autograd.grad(energy, q, 
        create_graph=True, retain_graph=True, grad_outputs=torch.ones_like(energy))[0]
\end{lstlisting}

\subsubsection{ALPS}
ALPS are modified from the code in \cite{ALPS} and it has the following basic modules: 1) Initial Encoder; 2) Physical Context Encoder; 3) State Encoder; 4) Neural Decoder. \\
\textbf{(1) Initial Encoder:}
\begin{lstlisting}
nn.Sequential(
    nn.Conv2d(time_steps, num_filters, kernel_size=5, stride=2, padding=(2, 2)),
    nn.BatchNorm2d(num_filters),
    nn.LeakyReLU(0.1),
    nn.Conv2d(num_filters, num_filters * 2, kernel_size=5, stride=2, padding=(2, 2)),
    nn.BatchNorm2d(num_filters * 2),
    nn.LeakyReLU(0.1),
    nn.Conv2d(num_filters * 2, num_filters, kernel_size=5, stride=2, padding=(2, 2)),
    nn.Tanh(),
    Flatten())
nn.Linear(num_filters*16, output_dim)
\end{lstlisting}
\textbf{(2) Physical Context Encoder:}
\begin{lstlisting}
nn.Sequential(
    nn.Linear(input_dim, 64),
    nn.ReLU(),
    nn.Linear(64, 64),
    nn.ReLU(),
    nn.Linear(64, 32)
    nn.ReLU(),
    nn.Linear(32, output_dim))
\end{lstlisting}
\textbf{(3) State Encoder:}
\begin{lstlisting}
class SpatialTemporalBlock(nn.Module):
    def __init__(self, t_in, t_out, n_in, n_out, last)
    self.conv = nn.Conv2d(n_in, n_out, kernel_size=5, stride=2, padding=(2,2))
    self.bn = nn.BatchNorm2d(n_out)
    self.act = nn.LeakyReLU(0.1)
    self.lin_t = nn.Linear(t_in, t_out)
nn.Sequential(
    SpatialTemporalBlock(time_steps, time_steps*2, 1, num_filters, False),
    SpatialTemporalBlock(time_steps*2, time_steps*2, num_filters, num_filters*2, False),
    SpatialTemporalBlock(time_steps*2, time_steps, num_filters*2, num_filters, True))
nn.Linear(num_filters*16, output_dim)
\end{lstlisting}
\textbf{(4) Neural Decoder:}
\begin{lstlisting}
nn.Sequential(
    nn.Linear(input_dim, 32),
    nn.ReLU(),
    nn.Linear(32, 32*4),
    nn.ReLU(),
    nn.Linear(32*4, 32*16),
    nn.ReLU(),
    nn.Linear(32*16, output_dim*output_dim))
\end{lstlisting}

\end{document}